\documentclass[aps,prd,reprint,superscriptaddress,floatfix]{revtex4-2}

\usepackage{xcolor}
\usepackage{graphicx} % Required for inserting images
\usepackage{amsmath}  % Essential for standard math typesetting
\usepackage{amssymb}
\usepackage{amsthm}   % Added for appendix proofs

\usepackage{placeins} % Provides the \FloatBarrier command to control float placement
\usepackage{siunitx} % Provides the \qty command for typesetting physical quantities with units

\usepackage{hyperref} % Adds clickable links to equations and references
\usepackage[capitalize]{cleveref} % Automatically handles contextual cross-referencing types (Eq., Fig., Sec.)
\begin{document}

\title{Learning Schwarzschild Geodesic Motion with Lagrangian Neural Networks}

\author{Sukru Caglar}
\email{sukru.caglar@std.bogazici.edu.tr}
\affiliation{Department of Physics, Bogazici University, Istanbul, Turkey}

\author{Sena Kalabalik}

\affiliation{Department of Physics, Bogazici University, Istanbul, Turkey}

\author{Abdullah Umut Hamzaogullari}

% List multiple affiliations one after the other for a single author
\affiliation{Department of Physics, Bogazici University, Istanbul, Turkey}
\affiliation{Department of Computer Engineering, Bogazici University, Istanbul, Turkey}

\author{Arkadas Ozakin}
\affiliation{Department of Physics, Bogazici University, Istanbul, Turkey}
\date{\today}

\begin{abstract}
Lagrangian Neural Networks (LNNs) learn dynamics from trajectory data by parameterizing a scalar Lagrangian and deriving accelerations through the Euler–Lagrange equation, providing a variational inductive bias that improves physical consistency over purely data-driven models. LNNs have recently been extended to general relativistic settings, leaving Schwarzschild geodesics as an open challenge. Training on complex relativistic systems is hindered by the diversity of the dynamics, as well as numerical instabilities and the difficulty of adequately sampling physically relevant, diverse orbital regimes. We introduce two coordinated improvements: a decoupled position–velocity normalization scheme that independently rescales each coordinate and its time derivatives, reweighting acceleration targets to balance contributions across all channels; and an energy-constrained log-uniformly sampled dataset generation strategy informed by the innermost stable circular orbit (ISCO) structure. With these improvements, our model accurately reproduces perihelion precession and relativistic time dilation, conserves energy and angular momentum to within a few percent even after completing multiple orbits around the central mass and predicts orbital trajectories across eight orbital scenarios in close, medium, and far-field regimes. These results establish that LNNs, augmented with physically motivated normalization and sampling, can reliably learn geodesic dynamics in curved spacetime. Our work also represents the first successful training of LNNs on Schwarzschild geodesics and offers a promising direction for data-driven modeling of complex relativistic systems.
\end{abstract}

\maketitle

\section{Introduction}
Learning the dynamics of a system directly from trajectory data has become a central goal in scientific machine learning \cite{chenNeuralOrdinaryDifferential2019}.
Among these approaches, Lagrangian Neural Networks (LNNs) \cite{cranmerLagrangianNeuralNetworks2020} are particularly appealing because they parameterize a scalar Lagrangian and derive equations of motion through Euler-Lagrange structure:
\begin{align}
    \ddot{\mathbf{q}} = (\nabla_{\dot{\mathbf{q}}} \nabla_{\dot{\mathbf{q}}}^{\top} \mathcal{L})^{-1} [\nabla_{\mathbf{q}} \mathcal{L} - (\nabla_{\mathbf{q}} \nabla_{\dot{\mathbf{q}}}^{\top} \mathcal{L}) \dot{\mathbf{q}}]
\end{align}
Rather than fitting accelerations directly, LNNs learn a scalar Lagrangian whose functional derivatives through the Euler–Lagrange operator produce the equations of motion. 
This variational framing imposes a geometric inductive bias: the mass matrix (velocity Hessian) encodes the system's effective metric, while the stationary-action principle ensures 
accelerations are consistent with an underlying scalar potential. As a result, LNNs tend to exhibit better physical fidelity, energy conservation, and out-of-distribution extrapolation than models that bypass this structure.

Our work directly builds on the stabilized LNN baseline of Hamzaogullari and Ozakin \cite{Hamzaogullari:2026}.
That framework introduced key Sylvester regularization and Lorentzian prior for mass matrix to stabilize the training in relativistic settings, successfully demonstrating geodesic learning in Anti-de Sitter ($AdS_4$) spacetime.

Schwarzschild geodesic motion is also a strong testbed for this regularized LNN structure. However, extending this baseline to the Schwarzschild spacetime exposes vulnerabilities due to substantial regime diversity \cite{Lim:2024mkb}. System contains near-circular bound orbits with subtle relativistic precession, 
higher-eccentricity trajectories with stronger radial variation, and trajectories near stability boundaries which are sensitive to small local error accumulation over long rollouts. The need to optimize for different regimes simultaneously makes the learning task more challenging. A successful model 
must therefore provide both low one-step prediction error for variety of regions and stable long-horizon integration with physically meaningful behavior.

We also observe additional failure modes tied to mixed-scale coordinates and derivatives, where input coordinates with vastly different magnitudes cause poorly conditioned optimization and loss domination by a subset of input channels.

A second challenge is data quality in six-dimensional phase space.
Uniform sampling of $(q,\dot{q})$ tends to overproduce plunge or escape states while under-sampling delicate bound-orbit regions, reducing data efficiency and degrading performance in the regimes most relevant for geodesic learning.

To address these issues, we introduce a coordinated set of improvements on top of the baseline framework \cite{Hamzaogullari:2026}: 
\begin{itemize}
    \label{sec:contributions}
    \item a physics-aware normalization strategy that decouples spatial and velocity scales and reweights acceleration targets.
    \item an energy-constrained dataset generation scheme informed by ISCO structure to control eccentricity coverage.
\end{itemize}

We also retain stabilization mechanisms introduced by \cite{Hamzaogullari:2026} that mitigate pathological Hessian behavior during training and rollout.

The remainder of this paper is structured as follows. In \cref{sec:background} we review the Lagrangian Neural Network framework and the Schwarzschild geodesic 
equations that define our learning target. \cref{sec:related} surveys related work in physics-informed learning and relativistic ML. 
\cref{sec:methods} details our two key methodological contributions and explains how each addresses the failure modes identified in the baseline approach. 
\cref{sec:experiments} presents a comprehensive experimental evaluation organized around training convergence (\cref{sec:norm_compare}), trajectory prediction accuracy close, medium, and far orbital regimes (\cref{sec:traj}), 
time-dilation fidelity (\cref{sec:time_dilation}), and structural analysis of the learned acceleration functions, Lagrangian, and mass matrix (\cref{sec:accel_check} and \cref{sec:hessian_chapter}). 
Finally, \cref{sec:conclusion} summarizes our findings and outlines directions for future work.

\section{Background}\label{sec:background}

\subsection{Lagrangian Neural Networks:}

\subsubsection{Acceleration targets from Blackbox Lagrangian:}
Euler-Lagrange equation, in its general form, provides us a system of differential equations to solve for $q(t)$ when the function $\mathcal{L}_\theta(q,\dot{q})$ is given. 
Since we assume $\mathcal{L}_\theta$ is a black box function\cite{cranmerLagrangianNeuralNetworks2020}, we rearrange the vectorized version of the Euler-Lagrange equation to find an expression for $\ddot{q}$ 
\begin{align}
    \frac{d}{d\tau} \left( \frac{\partial \mathcal{L}_\theta}{\partial \dot{q}_i} \right) = \frac{\partial \mathcal{L}_\theta}{\partial q_i} &\implies \frac{d}{d\tau} \nabla_{\dot{\mathbf{q}}}\mathcal{L}_\theta = \nabla_{\mathbf{q}}\mathcal{L}_\theta 
\end{align}
where $\dot{q}=\frac{dq}{d\tau}$. Distributing the time derivatives gives the final relation for $\ddot{q}$ as:
\begin{align}
    \label{eq:thegodeq}
   \ddot{\mathbf{q}} = (\nabla_{\dot{\mathbf{q}}} \nabla_{\dot{\mathbf{q}}}^{\top} \mathcal{L}_\theta)^{-1} [\nabla_{\mathbf{q}} \mathcal{L}_\theta - (\nabla_{\mathbf{q}} \nabla_{\dot{\mathbf{q}}}^{\top} \mathcal{L}_\theta) \dot{\mathbf{q}}]
\end{align}

The idea behind an LNN is to use \eqref{eq:thegodeq} as a way to 
deduce accelerations from a $\mathcal{L}_{\theta}(q,\dot{q})$ where we do not know the exact form of $\mathcal{L}_{\theta}$ but can apply the gradient operators ($\nabla_q,\nabla_{\dot{q}}$) using the gradients from our network.

\subsubsection{Training objective and loss}
\label{sec:lnn_loss}
Given a dataset of $N$ state-acceleration pairs $\{(q_i, \dot{q}_i), \ddot{q}_{i}\}$, we train the network by minimizing the mean absolute error between the predicted and ground-truth accelerations:
\begin{align}
    \label{eq:loss}
    L_{train} = \frac{1}{N} \sum_{i=1}^N \left|\ddot{q}_\theta(q_i, \dot{q}_i) - \ddot{q}_{i} \right|,
\end{align}
where $\ddot{q}_\theta$ is computed from the learned Lagrangian $\mathcal{L}_\theta$ via Eq.~\eqref{eq:thegodeq}.
The ground-truth targets $\ddot{q}$ can be obtained from Eq.~\eqref{eq:lag} by solving Euler-Lagrange Equations of the system.

\subsubsection{Lagrangian Parameterization with Lorentzian Structural Prior and Sylvester Regularization:}
\label{sec:lorentzian_prior}

Applying physics-aware neural networks to general relativity presents unique challenges due to the Lorentzian signature of the spacetime metric and the extreme nonlinearities near massive bodies.
The Stabilized LNN framework developed by Hamzaogullari and Ozakin \cite{Hamzaogullari:2026} introduces a Lorentzian Sylvester regularization scheme that enforces and maintains valid physical signatures on the learned mass matrix.
To incorporate this prior, we decompose the learned Lagrangian $\mathcal{L}_\theta(q,\dot q)$ into a fixed kinetic term with Lorentzian signature and a learnable perturbation:
\begin{equation}
    \mathcal{L}_\theta(q,\dot q) = \Bigl(-\dot q_0^2 + \sum_{i=1}^{n-1} \dot q_i^2\Bigr) \;+\; \mathrm{MLP}_\theta(q,\dot q)
    \label{eq:combined_decomposition}
\end{equation}
where $n = 3$ is the number of generalized coordinates, and $\dot q_0 \equiv \dot t$ is the time-velocity component.
The MLP perturbation is implemented as a fully connected network whose architecture hyperparameters are detailed in \cref{tab:architecture}.
By introducing a Lorentzian base term in\eqref{eq:combined_decomposition}, we bias the velocity Hessian toward the
desired Lorentzian signature at initialization, provided that the initial MLP contribution to the Hessian is sufficiently small.

In general relativity, the spacetime metric $g_{\mu\nu}$ has Lorentzian signature $(-,+,+)$. \cite{carroll2019spacetime}
Therefore, for Schwarzschild geodesics, the Lagrangian $\mathcal{L} = \frac12 g_{\mu\nu}\dot x^\mu\dot x^\nu$ must satisfy $\partial^2\mathcal{L}/\partial \dot t^2 < 0$ and the spatial velocity Hessian must be positive-definite.
Any learned Lagrangian that violates this signature is penalized by adding the following Sylvester criterion regularization term  \cite{Hamzaogullari:2026} to the training loss:
\begin{align}
L_{\text{relativistic}}
&= L_{\text{train}}
+\lambda\Biggl(
\frac{\operatorname{ReLU}(M_{00})}{c_0+\epsilon}
\nonumber\\
&\qquad
+\sum_{k=1}^{n-1}
\frac{
\operatorname{ReLU}\!\left(
-\det\!\left(M_k^{\mathrm{spatial}}\right)
\right)
}{
s_k^{\mathrm{spatial}}+\epsilon
}
\Biggr).
\label{eq:relativistic_loss}
\end{align}

\subsection{Schwarzschild geodesics:}

\subsubsection{Geodesic motion in Schwarzschild spacetime}
The Schwarzschild solution to vacuum Einstein's equation only took a few months to be found \cite{Schwarzschild1916,schwarzschild1999gravitational} after the introduction of Einstein's theory of general relativity in 1915, 
with Karl Schwarzschild deriving the solution in a short period while serving on the Russian front during World War I. Schwarzschild metric describes spherically symmetric solution for Einstein's equation in vacuum and 
produces the following metric:
\begin{equation}
    \label{eq:metric}
    \begin{aligned}
        ds^2 = &-\left(1 - \frac{r_s}{r}\right)dt^2 + \left(1 - \frac{r_s}{r}\right)^{-1}dr^2 \\
               &+ r^2 d\theta^2 + r^2\sin^2\theta d\phi^2
    \end{aligned}
\end{equation}

where $r_s=2M$ in simplified $G=c=1$ frame. Due to symmetries of this metric, we can also define two conserved quantities for a particle moving in this spacetime as specific energy and specific angular momentum which are given as:
\begin{align}
    \label{eq:killing1}
    E &= \left(1-\frac{r_s}{r}\right)\dot{t} \\
    \label{eq:killing2}
    L &= r^2\dot{\phi}
\end{align}
Derivations for above equations can  be found at Appendix \ref{sec:killing}.

Geodesic motion under this metric can be obtained by solving the geodesic equation:
\begin{align}
    \label{eq:geodesic}
    \frac{d^2x^\mu}{d\tau^2} + \Gamma^\mu_{\nu\lambda}\frac{dx^\nu}{d\tau}\frac{dx^\lambda}{d\tau} = 0
\end{align}

Notice again that time derivatives are taken with respect to an affine proper time parameter $\tau$. System can also be represented equally well by using the Euler-Lagrange  equation and the following Lagrangian:
\begin{gather}
    \label{eq:lag}
    \mathcal{L} = \frac{1}{2} g_{\mu\nu} \dot{x}^\mu \dot{x}^\nu 
\end{gather}
\cref{eq:lag} also represents the kinetic term of a particle moving at some geodesic path inside a curved space.

In this work, we examine the system on the equatorial plane where $\theta=\pi/2$ and $\dot{\theta}=0$.
This substitution transforms Eq.~\eqref{eq:lag} into:
\begin{equation}
    \label{eq:lagrangian}
    \mathcal{L} = \frac{1}{2}\left[-\left(1-\frac{r_s}{r}\right)\dot{t}^2+\left(1-\frac{r_s}{r}\right)^{-1}\dot{r}^2+r^2\dot{\phi}^2\right]
\end{equation}
This represents the expected form of the Lagrangian of the system.

The acceleration equations can be equivalently obtained either from the Euler--Lagrange equations or directly from the geodesic equations of motion~\eqref{eq:geodesic} for a timelike particle ($u^\mu u_\mu = -1$). Expressed as functions of the 6D input state $(q,\dot{q})$, the equations of motion on the equatorial plane are:
\begin{align}
    \label{eq:accel}
    \ddot r &= -\frac{M}{r^2} + (r-3M)\dot\phi^2, \\
    \ddot\phi &= -\frac{2\dot r\dot\phi}{r}, \\
    \label{eq:t_acell}
    \ddot t &= -\frac{r_s\dot t\dot r}{r(r-r_s)} = -\frac{2M\dot t\dot r}{r(r-2M)}.
\end{align}

\subsubsection{Mercury's Perihelion Precession}

One of the earliest and most celebrated confirmations of general relativity was its resolution of the anomalous perihelion precession of Mercury.
Newtonian mechanics, including the perturbative effects of other planets, predicts a precession of approximately \qty{5557}{\arcsecond} per century, yet observations yielded approximately \qty{5600}{\arcsecond} per century---an excess of roughly \qty{43}{\arcsecond} that remained unexplained until Einstein's general theory of relativity~\cite{carroll2019spacetime}.
In the weak-field limit of Schwarzschild geodesics, the relativistic correction to the effective potential naturally produces this additional \qty{43}{\arcsecond} per century, making Mercury's orbit an iconic benchmark for any computational approach to geodesic motion.
While our present work focuses on the strong-field regime ($r \sim 50$--$300$ in units of $r_s/2$) rather than the solar-system weak field, Mercury's precession provides the historical and physical motivation for learning Schwarzschild dynamics from data. 
For a comparison between our regime and the solar-system weak field, refer to the discussion in \cref{sec:extrapolation}.

\section{Related Work}\label{sec:related}

The endeavor to extract analytical physical laws directly from observational data has been significantly advanced by the integration of geometric priors into deep learning.
Our approach builds upon several intersecting domains: physics-informed neural networks, techniques for overcoming optimization instabilities in these architectures, and the application of machine learning to general relativistic dynamics.

\subsection{Structure-Preserving Neural Dynamics}

Neural ordinary differential equations (Neural ODEs) parameterize a continuous-time vector field with a neural network and evaluate the resulting dynamics using a numerical 
ODE solver \cite{chenNeuralOrdinaryDifferential2019}. Although this provides a flexible framework for learning dynamical systems, the generic formulation does not impose the 
mechanical structure or conserved quantities of the underlying physical system. Hamiltonian Neural Networks (HNNs) address this limitation by parameterizing a scalar Hamiltonian
and deriving the phase-space vector field through Hamilton's equations \cite{greydanusHamiltonianNeuralNetworks2019}. This construction introduces a conservation-oriented inductive bias, 
but requires the input variables to form canonical coordinate--momentum pairs.

Lagrangian approaches provide an alternative formulation in terms of generalized coordinates and velocities. Deep Lagrangian Networks (DeLaNs) impose a 
kinetic--potential decomposition with a configuration-dependent positive-definite mass matrix, making them particularly suitable for classical rigid-body 
systems and control applications \cite{Lutter:2019ovq}. Black-box Lagrangian Neural Networks (LNNs), by contrast, directly parameterize an unrestricted scalar 
Lagrangian and obtain accelerations through the Euler--Lagrange equations \cite{cranmerLagrangianNeuralNetworks2020}. They therefore do not require predefined 
canonical momenta and can represent nonquadratic velocity dependence, as demonstrated in the original LNN work for a relativistic particle. These structural 
priors promote physically consistent dynamics, although exact conservation during rollout is not guaranteed when the learned scalar function and numerical 
integrator are approximate.

Related work has also combined structured neural representations with symbolic regression to recover interpretable analytical expressions from trained models 
\cite{cranmerDiscoveringSymbolicModels2020}. Such symbolic extraction is a separate post-training procedure rather than an inherent guarantee of the LNN parameterization. 
In the present work, we adopt the black-box LNN formulation together with the stabilized Lorentzian framework of Hamzaogullari and Ozakin \cite{Hamzaogullari:2026}, 
since Schwarzschild geodesics are naturally expressed using generalized spacetime coordinates and an indefinite velocity Hessian.

\subsection{Stabilizing Lagrangian Neural Networks}

Black-box LNNs obtain accelerations by solving a linear system involving the velocity Hessian
($H_{\dot q\dot q}=\nabla_{\dot q}\nabla_{\dot q}^{\top}\mathcal{L}_{\theta}$).Consequently, a singular or poorly conditioned learned Hessian can amplify 
approximation errors, destabilize gradient-based optimization, and produce rapidly diverging trajectory rollouts \cite{cranmerLagrangianNeuralNetworks2020}. 
This conditioning problem is specific to the Euler--Lagrange forward map and is distinct from the first-derivative construction used in standard HNNs.

Previous approaches control this problem by imposing different forms of prior structure. DeLaNs explicitly parameterize a positive-definite mass matrix and 
a potential-energy function, improving mechanical plausibility at the cost of restricting the admissible Lagrangian form \cite{Lutter:2019ovq}. Finzi et al. 
instead embed systems in redundant Cartesian coordinates and enforce known holonomic constraints using Lagrange multipliers \cite{finziSimplifyingHamiltonianLagrangian2020}. 
This avoids coordinate singularities associated with some generalized-coordinate representations and can substantially simplify the learned functions, but assumes 
that the relevant Cartesian embedding and constraints are known in advance.

The closest methodological precursor to our work is the stabilized LNN framework of Hamzaogullari and Ozakin \cite{Hamzaogullari:2026}. 
That framework combines velocity-Hessian signature regularization, activation functions suited to higher-order differentiation, 
and physics-aware coordinate scaling. Its Lorentzian extension penalizes departures from the required indefinite Hessian 
signature and was demonstrated on geodesic motion in ($AdS_4$) spacetime. We retain this Lorentzian stabilization strategy but address
an additional difficulty that arises in Schwarzschild dynamics: the coordinates, velocities, and acceleration components occupy substantially 
different numerical scales. Rather than introducing explicit Cartesian constraints or enlarging the input representation, we independently 
normalize the position and velocity channels and consistently rescale and weight the acceleration targets. The network consequently learns 
a latent scalar directly as a function of the normalized variables that acts as effective Lagrangian while preserving the standard algebraic form of the Euler--Lagrange forward model.

\subsection{Relativistic Machine Learning and Schwarzschild Phase-Space Sampling}

Physics-informed neural networks (PINNs) incorporate known differential equations into the training objective by penalizing their residuals at 
sampled collocation points \cite{raissiPhysicsInformedNeural2019}. In relativistic applications, PINNs have been used to solve the Teukolsky 
equation and calculate black-hole quasinormal modes \cite{lunaSolvingTeukolsky2023}, while automatic-differentiation-based neural solvers have 
recovered Schwarzschild and charged Schwarzschild solutions from the Einstein field equations \cite{liSolvingEinstein2023}. More recently, 
Lorentzian physics-informed architectures have been used to recover Schwarzschild geometry by combining the vacuum Einstein equations with 
geometric and symmetry constraints \cite{gherardiniBlackHoleBlack2026}. These approaches solve or constrain explicitly specified relativistic field equations. 
They therefore differ from trajectory-based Lagrangian learning, in which a scalar dynamical function is inferred from state--acceleration data and 
the equations of motion are generated through the Euler--Lagrange operator.

The closest trajectory-based relativistic study is the stabilized LNN framework of Hamzaogullari and Ozakin, which demonstrated the learning of 
geodesic dynamics in ($AdS_4$) spacetime \cite{Hamzaogullari:2026}. Extending this approach to bound Schwarzschild motion introduces an additional
data-selection problem because the accessible orbital regimes are structured by the conserved specific energy (E), specific angular momentum (L), 
and the Schwarzschild effective potential \cite{carroll2019spacetime,Lim:2024mkb}. Naive sampling of coordinates and velocities can consequently 
overrepresent plunge or escape configurations while providing insufficient coverage of the nearly circular and eccentric bound states needed to
learn long-term orbital dynamics.

To address this problem, we introduce an ISCO threshold informed, energy-filtered velocity-sampling strategy designed to improve coverage of the bound-orbit phase space. 
The full sampling procedure, physical acceptance criteria, and resulting datasets are described in \cref{sec:sampling}.

\section{Methods}\label{sec:methods}

We retain the Lagrangian parameterization, Lorentzian initialization, and Sylvester regularization of the stabilized LNN described in \cref{sec:lorentzian_prior}. We will refer to this structure as "baseline" for the remainder of this work.
As a result of the Lorentzian prior, Sylvester Criterion regularization follows the methodology for the relativistic systems described by \cite{Hamzaogullari:2026} and \cref{sec:lorentzian_prior} to penalize violations from the Lorentzian signature.
We introduce two modifications to address the additional difficulties encountered in Schwarzschild geodesic learning: decoupled normalization of position and velocity channels, together with per-channel acceleration weighting; and an ISCO threshold informed, energy-filtered velocity-sampling strategy. The first improves the numerical conditioning of the single-step acceleration objective, while the second concentrates the training data on the orbital regimes relevant to long-horizon bound motion.

\subsection{Normalization}\label{sec:normal}
\subsubsection{Decoupled normalization of positions and velocities}

It is clear that $\mathcal{L}$ will be a function of positions and velocities.
Accordingly, we choose our input and targets states to be:

\begin{align}
    \text{inputs} &\implies (q,\dot{q})=\langle t,r,\phi,\dot{t},\dot{r},\dot{\phi}\rangle\\
    \text{targets} &\implies (\ddot{q})=\langle \ddot{t},\ddot{r},\ddot{\phi}\rangle
\end{align}

The baseline architecture utilizes an independent physical scaling method, normalizing each coordinate by its maximum absolute value in the training dataset and imposing the same scaling to the velocities.
While effective for mechanically homogeneous systems like the double pendulum, this coupled scaling inherently assumes that the size of each spatial dimension is comparable in magnitude to their velocity.
When this constraint is not satisfied, signals from each position and velocity group become unbalanced.

For systems with multiple coordinates spanning different physical scales, such as Schwarzschild geodesics, where $r \sim \num{1e2}$ while $\dot{\phi} \sim \num{1e-3}$, bringing input channels to comparable numerical ranges is a standard preprocessing step that can improve the conditioning and convergence of gradient-based neural-network training \cite{lecun1998efficient,sola1997importance}. The acceleration components likewise have different characteristic scales; therefore, we measure their prediction errors relative to per-channel mean magnitudes so that the loss contribution to the MAE from each channel reflects comparable relative accuracy coordinates. This motivates the decoupled input scaling and scale-normalized acceleration loss introduced below.

To balance the numerical ranges of input channels, we implement a two-phase decoupled position-velocity normalization that applies a scale vector $\mathbf{S}$ to the position inputs and an effective velocity scale $\mathbf{S}/\mathbf{T}$ with components $S_i/T_i$, where $T_i$ denotes the characteristic time scale factor for coordinate $i$. Both scales represent $n$-dimensional vectors that independently normalize input channels in the $2n$-dimensional input state $(q,\dot{q})$ as:
\begin{gather*}
    \tilde{q}_i=S_i q_i \\
    \tilde{\dot{q}}_i=\frac{S_i}{T_i}\dot{q}_i
\end{gather*}

This substitution naturally scales the proper-time differential $d\tau$ as $d\tilde{\tau}_i = T_i d\tau$, so that normalized target accelerations are given by:
\begin{equation}
    \tilde{\ddot{q}}_i=\frac{S_i}{T_i^2}\ddot{q}_i
\end{equation}
This resulting scale is applied to training targets $\ddot{q}$ to preserve the physical connection between velocities and accelerations.

When the $(q,\dot{q})$ and $(\ddot{q})$ states are scaled accordingly, the imbalance between target accelerations are greatly reduced. To further reduce the remaining numerical imbalances and keep a balanced attention on each dimension, we apply scale-normalization to each MAE loss channel which inversely multiplies the magnitudes of the mean of each acceleration target in each dimension to make the numerical contributions to the base loss be comparable all channels.

\begin{equation}
L_{train}=\frac{1}{Nn}\sum_{k=1}^{N}\sum_{i=1}^{n}w_i\left|\tilde{\ddot q}_{\theta,i}\bigl(\tilde q_k,\tilde{\dot q}_k\bigr)
-\tilde{\ddot q}_{k,i}\right|
\label{eq:weighted_acc_loss}
\end{equation}
with:
\begin{equation}
m_i:=\frac{1}{N}\sum_{n=1}^{N}\left|\tilde{\ddot q}_{n,i}\right|,
\qquad w_i:=\frac{1}{m_i}
\label{eq:raw_acc_weights}
\end{equation}

Decoupled-scaling and loss normalization processes described above has greatly improved the training performance and masked both the imbalance in magnitudes dimensions and the imbalance between time derivatives of the dimensions.

%\subsubsection{diferentiation skewedness:}
%We must note that applying different scales to the velocities and position coordinates has a mathematical cost of altering the proper time derivative $d\tilde{t}$ with a vector whose components are different dimensions.
%As a result, the tensor components in the original euler lagrange equation suggested in the baseline work for infering the accelerations are no longer corretly representing the system with scaled time derivatives.
%This affect and how the structure of our metric based, velocity-decoupled mechanics of the shcwarzschild geodesic motion avoids it is explained in the Appendix.\ref{sec:norm}.
\subsubsection{Effect of decoupled normalization on the variational interpretation of the learned Lagrangian}

Applying independent scales $S_i$ and $T_i$ to each coordinate means the normalized time derivatives $\tilde{\dot q}_i$ measure evolution against different clock rates.
One might then worry that the Euler-Lagrange equation acquires cross-coordinate distortion factors in double derivatives in velocities, invalidating the standard acceleration formula of Eq.~\eqref{eq:thegodeq} in normalized space. A full derivation of the resulting distortion is given in \cref{sec:norm}, where we show explicitly that a rescaled physical Lagrangian would induce $T_j/T_i$ weighting factors in the acceleration equation.

Moreover, since the velocity scales are selected independently, $\tilde{\dot q}_i$ should not generally be interpreted as the derivative of $\tilde{q}_i$
with respect to a single normalized evolution parameter. The decoupled transformation is therefore not a tangent lift of a configuration-space coordinate transformation.
the learned function $\mathcal L_\theta(\tilde q,\tilde{\dot q})$
should be interpreted as a latent generating scalar for the normalized acceleration map, rather than as the physical Schwarzschild Lagrangian with only its arguments rescaled.
The model computes normalized accelerations through the Euler–Lagrange structured acceleration operator $\tilde{\ddot q} = \tilde H^{-1}[\tilde g - \tilde C \tilde{\dot q}]$, (refer to \cref{sec:norm} for matrix definitions here), and physical accelerations are recovered via the inverse scaling $\ddot q_i = (T_i^2/S_i)\,\tilde{\ddot q}_i$.

A result of this design is that the learned latent scalar $\mathcal{L}_\theta$ will generally differ in functional form from the physical Lagrangian since it is a latent space representation and not constrained to be a simple rescaling of $L_{\rm phys}$.
What this freedom implies for physical interpretability, is examined quantitatively in \cref{sec:hessian_chapter} and \cref{sec:norm}.

\subsection{Dataset Generation}\label{sec:sampling}
\subsubsection{Bound Training Data:}

One other difficulty of learning Schwarzschild geodesics is the sampling $q,\dot{q}$ from the phase space consistently to focus on the desired motion in the geodesics.
During the analysis of our baseline approach for dataset generation which samples randomly for an independent interval for each coordinate in  $<t,r,\phi,\dot{r},\dot{\phi}>$ and determines $\dot{t}$ according to \cref{eq:timelike_cons} to create input state $<t,r,\phi,\dot{t},\dot{r},\dot{\phi}>$, we saw that sample points with certain combinations of position and velocities results in plunging or escaping orbits when left to evolve.
Including this type of motions in the training regime highly increases the necessary number of combinations of ranges for each coordinate which we summarize as the phase-space of our dataset.

Moreover, learning plunge or escape orbits is more challenging in terms of normalization since it introduces outliers to the $(q,\dot{q})$ sample-space.
Therefore we introduced a more robust energy constrained, velocity sampling algorithm for bound orbits with velocities that covers a range of eccentricities for a given radius interval. The details of our sampling algorithm is given below. 

\subsubsection{Sampling \texorpdfstring{$\dot{r}, \dot{\phi}$}{r-dot, phi-dot}}

The overarching objective of machine learning models like LNN is to extract dynamical laws directly from observational data when the only known information is the real world observation.
Naturally, ideal justification for the success of such model would be an experiment based on the real-world data and for the cases where this is not possible, with a synthetic data that resembles the real-world data as best as possible.

In stellar-orbit studies of gravitational system with a spherical central mass and more specifically massive black holes, the direct measurements typically consist of sky-projected astrometric positions and spectroscopic redshifts or line-of-sight velocities, 
whereas complete trajectories and conserved quantities such as orbital energy are inferred through dynamical modeling \cite{do2019redshift,gravity2020precession}. Motivated by this distinction, 
we generate candidate states in position-velocity space with boundedness remaining as a strong observational constraint rather than prescribing an empirical distribution over orbital energy which would algorithmically be more straightforward with our energy constraints that we will explain in more detail shortly. 

To implement this, we sample the initial positions $(r, \phi)$ and velocities $(\dot{r}, \dot{\phi})$ directly from predefined boundaries that represent our observational window.
The radial coordinate $r$ is sampled log-uniformly to ensure sufficient coverage of strong field regions which are more informative for the network, while the velocities $\dot{r}$ and $\dot{\phi}$ are sampled from independent distributions.

For the learning task at hand, we choose the velocity distributions to be uniform. However, depending on the desired system dynamics, these velocity distribution may be normal to better approximate a relaxed cluster of orbiting bodies.
Once a candidate state $(r, \phi, \dot{r}, \dot{\phi})$ is drawn, we enforce the timelike normalization constraint for a massive particle ($u^\mu u_\mu = -1$) to compute $\dot{t}$, to fully define the state.
In the Schwarzschild metric, this invariant interval is expressed as:
\begin{equation}
    \label{eq:timelike_cons}
    -1 = -\left(1-\frac{r_s}{r}\right)\dot{t}^2 + \left(1-\frac{r_s}{r}\right)^{-1}\dot{r}^2 + r^2\dot{\phi}^2
\end{equation}

Or if we arrange for $\dot{t}^2$:
\begin{equation}
    \label{eq:fourvelocity}
    \dot{t}^2 = \frac{1 + \left(1-\frac{r_s}{r}\right)^{-1}\dot{r}^2 + r^2\dot{\phi}^2}{1-\frac{r_s}{r}}
\end{equation}

For valid physical states, we take the positive root to obtain $\dot{t}$.
With the full phase space state $(t, r, \phi, \dot{t}, \dot{r}, \dot{\phi})$ constructed, we compute the conserved quantities $E$ and $L$ using the definitions of specific energy \eqref{eq:killing1} and specific angular momentum \eqref{eq:killing2} in this spacetime.

Finally, we apply a rejection sampling mechanism based on the physical criteria established in the Appendix.\ref{sec:dat}.
A sampled state is only accepted into the training dataset if it satisfies both the innermost stable circular orbit threshold and the boundedness threshold:
\begin{gather}
    |L|
\ge L_{ISCO} = \sqrt{3} r_s \\
    E \le E_{max}
\end{gather}
where $E_{max} \approx \num{1}$ to include mostly but not necessarily bounded samples.
This rejection-based methodology is not present in baseline sampling method and ensures that our dataset is biased toward bound-compatible orbital
states while a controlled fraction of near-threshold configurations is maintained, providing a kinematic distribution that more closely resembles observable multi-body systems while performing better at our learning task.
For a more detailed discussion of the sampling algorithm and its effects on the training dataset distribution, refer to \cref{sec:setup}.

\section{Experiments}\label{sec:experiments} 

We present the setup and architectural parameters for our training in \cref{sec:setup} which mainly follows the methodology described in the previous section. 
Here we will assess the performance of the model in terms of  trajectory prediction fidelity.

We first assess the impact of our proposed normalization scheme, both in terms of loss convergence and trajectory prediction fidelity in \cref{sec:norm_compare}.

With the established methodology, we evaluate trajectory accuracy tests three orbital regimes (close, medium and far) where we benchmark predicted 
orbits against analytical geodesic solutions \cref{sec:traj}.

We complement the spatial analysis with an investigation of time dilation by analyzing the predicted time-velocities from the model all regions
in \cref{sec:time_dilation}, verifying that the relativistic coupling between the temporal and spatial coordinates is captured.

Finally, we probe the learned acceleration, Lagrangian and Hessian functions directly by treating pairs of input coordinates as independent variables and comparing the 
resulting acceleration surfaces to their analytical counterparts, providing a global view of the model's functional agreement with the true dynamics \cref{sec:accel_check}.

Throughout this work, we use geometric units with $G=c=1$, and set the
black-hole mass to $M=1$ in the numerical calculations. Length and time
quantities are therefore reported relative to $M$. In particular, the
trajectory evolution parameter is the dimensionless proper time
$\tau/M$. Dots denote differentiation with respect to the physical
proper time $\tau$.
\FloatBarrier

\subsection{Effects of Normalization on Training Performance}\label{sec:norm_compare}

\subsubsection{Loss Behavior:}
\begin{figure}[htbp]
    \centering
    \includegraphics[width=\columnwidth]{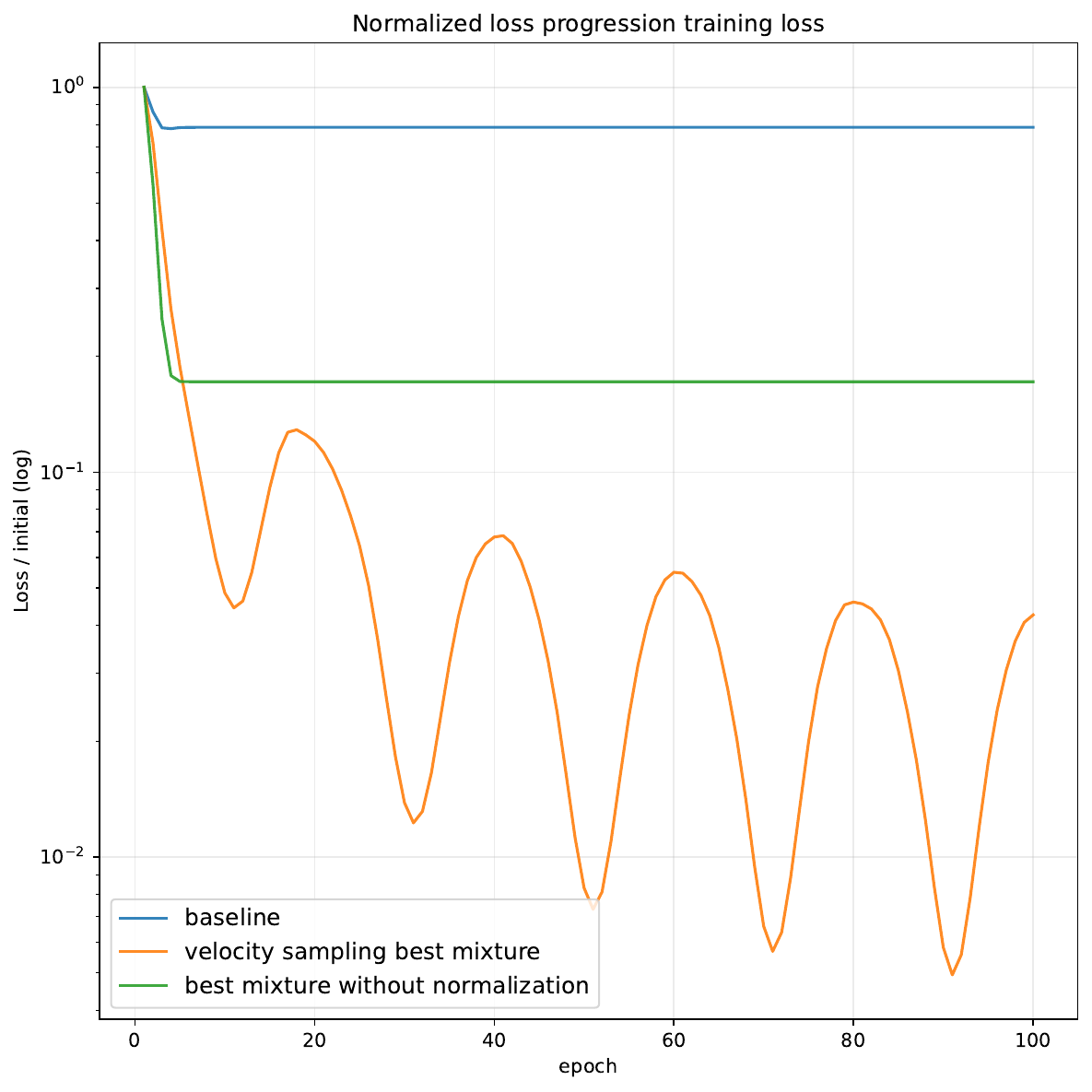}
    \caption{Progression of loss graph for not normalized and normalized models.
Y axis represents the ratio between the loss value in current epoch and the initial loss.}
    \label{fig:losses}
\end{figure}

For each dataset outlined in the \cref{tab:dataset_params}, we tracked the training performance with and without the decoupled normalization.
Because the target scales and per-coordinate weights shifts the numerical value of the MAE loss from $\mathcal{O}(\num{1e-8})$ to $\mathcal{O}(\num{1})$, raw loss magnitudes of decoupled normalized and legacy normalized dataset trainings are not directly comparable.
We therefore compare the fractional training loss progression relative to the initial value of each run.
\begin{equation}
    \widehat{L}(e)
    =
    \frac{L(e)}{L(0)}
\end{equation}
 From \cref{fig:losses}, it is clear that the decoupled normalization scheme significantly improves convergence, while the legacy normalization fails to reduce the loss effectively.
 For a more detailed comparison of loss progressions, refer  to \cref{sec:setup}, \cref{fig:losses_all}. 

\subsubsection{Predictions with and without Decoupled Normalization:}

\begin{figure*}[htbp]
    \centering
    \includegraphics[width=\textwidth]{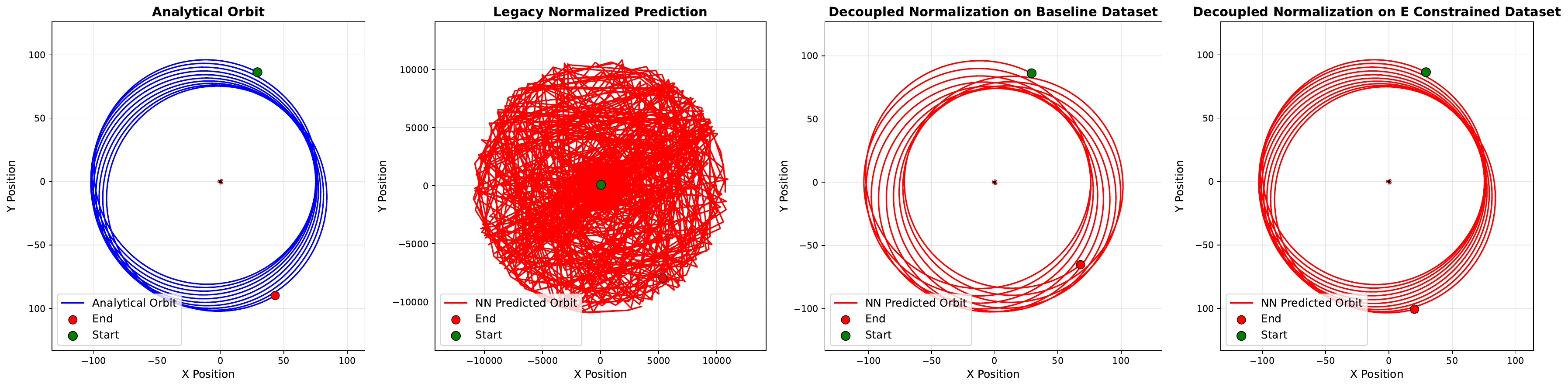}
    \caption{Effects of normalization and training dataset on the trajectory predictions.
            Analytical solution(left), legacy normalization training with balanced mixture dataset(middle left), baseline dataset training with normalization(middle right) and best mixture dataset training with decoupled normalization(right).Comparing the second and fourth panels isolates the effect of the normalization
            scheme on the mixed-velocity dataset. Comparing the third and fourth panels instead isolates the effect of dataset sampling while retaining the proposed normalization.}
    \label{fig:norm_effects}
\end{figure*}

Regardless of the dataset, it is clear from \cref{fig:losses} that loss values do not minimize effectively when the training is legacy normalized, while training with decoupled normalization shows convergence.
And the effects on the trajectory predictions from \cref{fig:norm_effects}, shows that the legacy normalization fails to extract information about the system when the signals from different dimensions are not balanced.

\cref{fig:norm_effects} also offers another result about the effects of the dataset choice when we compare baseline dataset training with mixed velocity dataset training where in this case, both uses decoupled normalization. 
The discrepancy in the predictions suggests that random state sampling remains insufficient to fully learn the dynamics of the system. (Also see \cref{tab:dataset_params} and \cref{fig:dists} for the distribution of the datasets)

\FloatBarrier
\subsection{Trajectory Accuracy}\label{sec:traj}

We train our model in the regime $r \in [\num{50}, \num{300}]$ with the non-plunging trajectory samples.
Despite this restriction, training set includes a variety of behaviors from near circular orbits to highly precessing ellipses or unbound/scattering trajecories.
We tested our model in 3 different regions of this boundaries with 8 different initial configurations with different amount of eccentricities.
For each initial condition, we solve the differential equation for x from learned and analytical solutions and let them evolve for $\tau/M=5\times10^4$, with time steps of  $\Delta\tau/M=10$.
Then the performance of the models is evaluated using the consistincy of the orbital paths between learned and analytical solutions, energy and angular momentum conservation error throughout the test time span and individual coordinate and velocity comparisons.

\subsubsection{Close Region:}
Close region for our training boundaries consists of $r \in [\num{50}, \num{90}]$ where the most dominant time dilation and precession effects are contained.
Due to its relatively strong GR effects this region has high importance in terms of benchmarking the learned dynamics of the system.
We experimented with orbits with 3 different eccentricities starting at $r=\num{60}$ with varying radial and angular speeds.

\begin{table}[htbp]
\caption{Initial conditions and orbital metrics for close orbit scenarios.}
\label{tab:close_orbit_metrics}
\begin{ruledtabular}
\resizebox{\columnwidth}{!}{%    
\begin{tabular}{lc c c c c c}
 & \textbf{$\dot{r}$($\times 10^{-1}$)} & \textbf{$\dot{\phi}$}($\times 10^{-3}$) & \textbf{$L$} & \textbf{$E$} & \textbf{$e$} & \textbf{$r_{min}$ - $r_{max}$} \\
\colrule
1 & $[\num{0.1}, \num{0.2}]$ & $[\num{2.2}, \num{2.4}]$ & \num{8.4632} & \num{0.9930} & \num{0.1902} & \numrange{57.51}{84.53} \\
2 & $[\num{0.1}, \num{0.2}]$ & $[\num{2.4}, \num{2.6}]$ & \num{9.1858} & \num{0.9948} & \num{0.3769} & \numrange{58.91}{130.17} \\
3 & $[\num{0.1}, \num{0.2}]$ & $[\num{2.7}, \num{2.8}]$ & \num{10.0129} & \num{0.9969} & \num{0.6285} & \numrange{59.40}{260.42} \\
\end{tabular}%
}
\end{ruledtabular}
\end{table}

Here we present one of the test cases from scenario 2. The rest of the scenario predictions can be found at \cref{sec:predictions}.

\begin{figure}[htbp]
    \centering
    \includegraphics[width=\columnwidth]{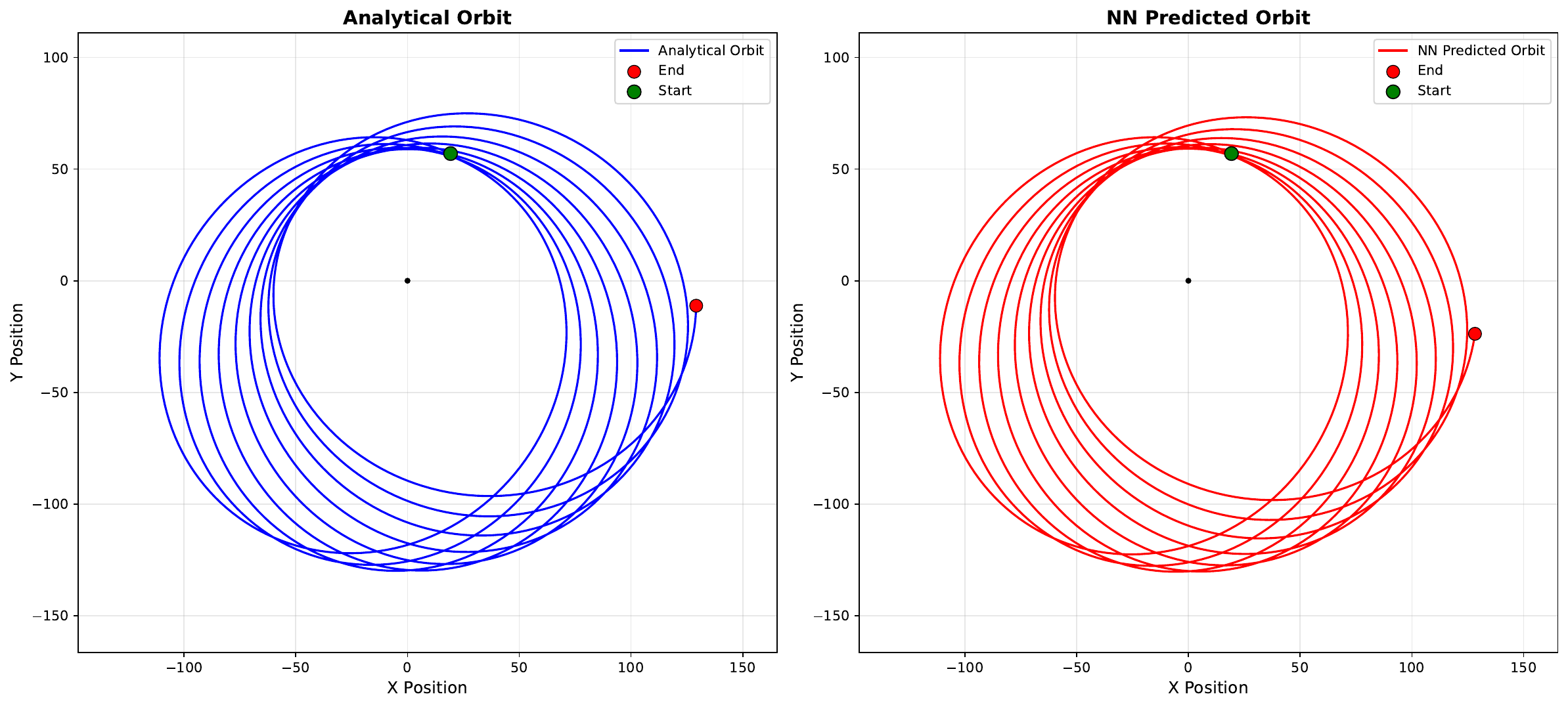}
    \caption{Side-by-side comparison for $e=\num{0.3769}$ test orbit.
Analytical to the left and model prediction to the right.}
    \label{fig:close_region_less_const3}
\end{figure}

\begin{figure}[htbp]
    \centering
    \includegraphics[width=\columnwidth]{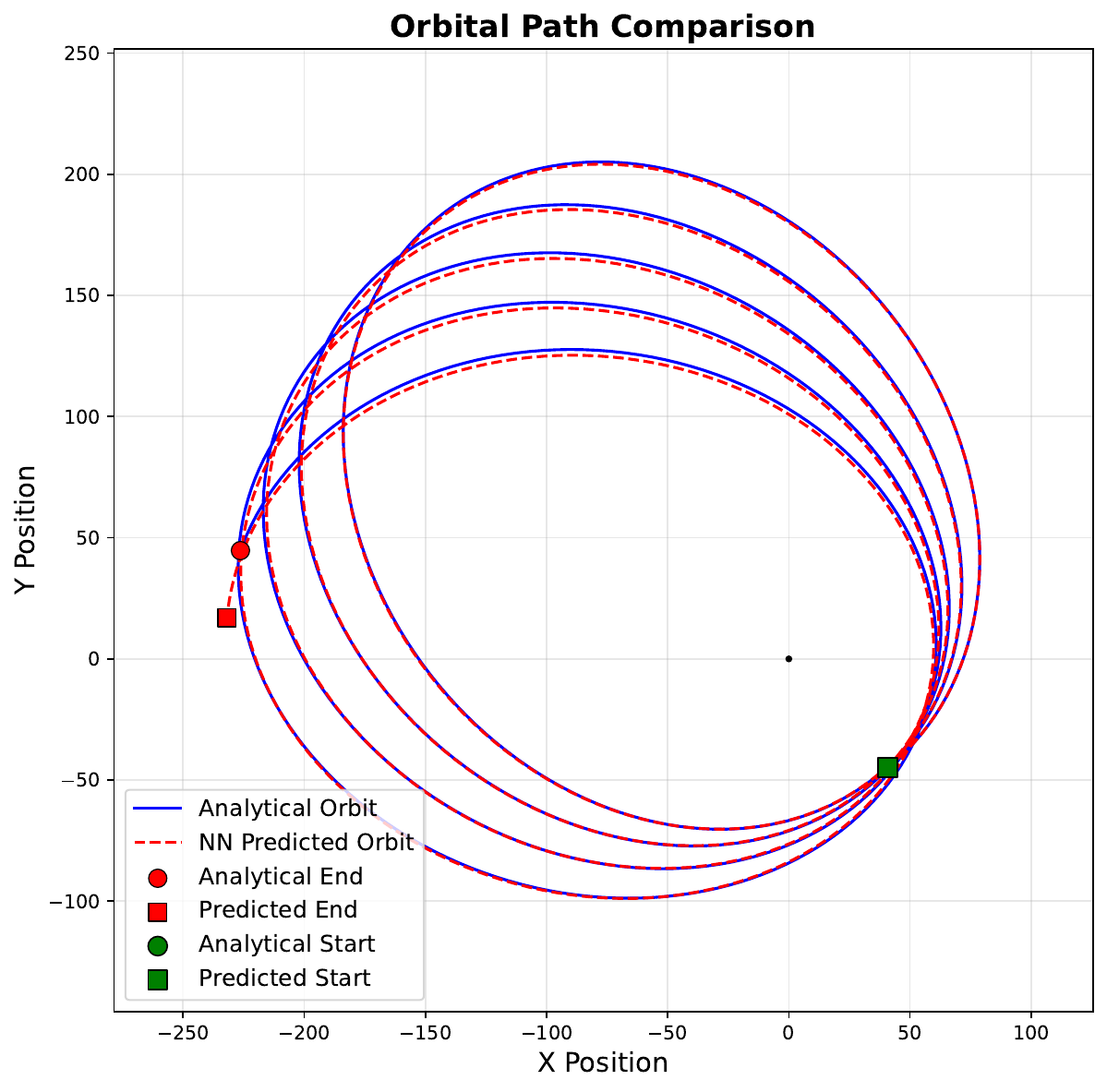}
    \caption{Overlay of analytic and predicted trajectories from scenario 3. The error accumulation in the precession is slight yet visible as the orbit completes new revolutions.}
    \label{fig:close_region_less_const3_combined}
\end{figure}

The accumulation of positional error due to stepwise numerical integration where output of each integration step serves as the input for the next, causing local deviations to propagate is visible in \cref{fig:close_region_less_const3_combined}.

\begin{figure}[htbp]
    \centering
    \includegraphics[width=\columnwidth]{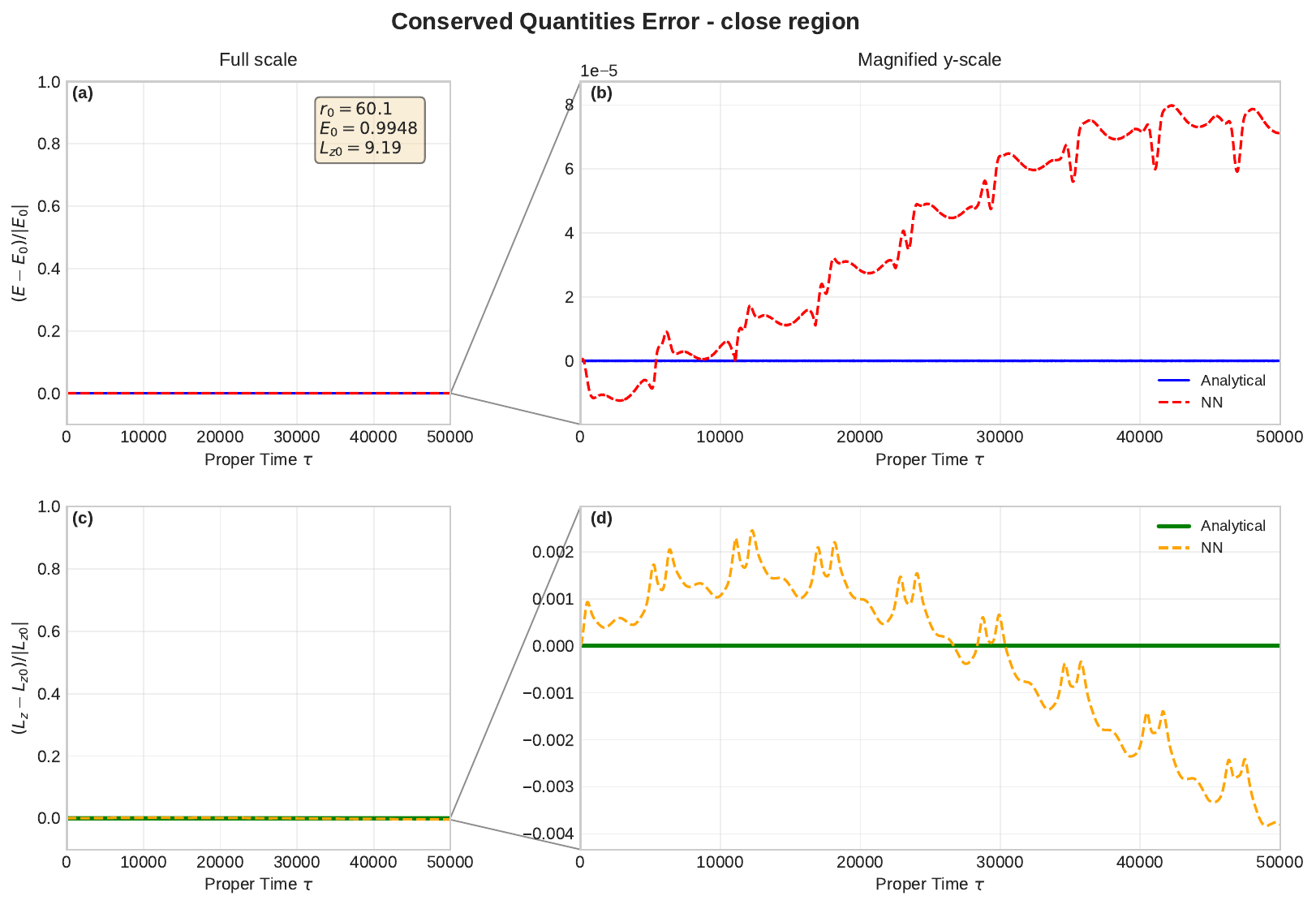}
    \caption{Energy conservation error (top) and angular 
momentum conservation error (bottom) for close region scenario 2. Analytical values for both quantities are computed from the Killing constants, which are conserved along the geodesic.}
    \label{fig:energy_cons_less_const3}
\end{figure}

We also present the energy and angular momentum (\cref{eq:killing1,eq:killing2}) conservation error for this scenario in \cref{fig:energy_cons_less_const3}.

\FloatBarrier
\subsubsection{Medium Region}
The medium region represents the mechanics within $r \in [\num{90}, \num{200}]$ units.
It has less dominant relativistic effects, yet it is still offers a wide range of orbital configurations, which, when eccentricities increase, still makes the precession phenomena visible.

We have 3 test scenarios starting at $r=\num{91}$ for this region, 2 of which have eccentricity $> \num{0.1}$ and almost a circular trajectory to stress test the stability of the predictions.

\begin{table}[htbp]
\caption{Initial conditions and orbital metrics for medium orbit scenarios.}
\label{tab:standard_orbit_metrics}
\begin{ruledtabular}
\resizebox{\columnwidth}{!}{%    
\begin{tabular}{lc c c c c c}
 & \textbf{$\dot{r}$($\times 10^{-1}$)} & \textbf{$\dot{\phi}(\times 10^{-3})$} & \textbf{$L$} & \textbf{$E$} & \textbf{$e$} & \textbf{$r_{min}$ - $r_{max}$} \\
\colrule
1 & $[\num{0.1}, \num{0.2}]$ & $[\num{1.0}, \num{1.2}]$ & \num{9.4722} & \num{0.9944} & \num{0.1533} & \numrange{75.08}{102.27} \\
2 & $[\num{0.1}, \num{0.2}]$ & $[\num{1.25}, \num{1.35}]$ & \num{11.6392} & \num{0.9968} & \num{0.3945} & \numrange{94.83}{218.39} \\
3 & $[\num{0.0}, \num{0.0}]$ & $[\num{1.17}, \num{1.19}]$ & \num{9.6338} & \num{0.9945} & \num{0.0101} & \numrange{88.81}{90.62} \\
\end{tabular}%
}
\end{ruledtabular}
\end{table}

Predictions for scenario 1 and 2 are equally good (\cref{fig:traj_accuracy}) as the one presented 
in the close region scenario 2. A more informative investigation would be looking at the scenario 3 where the test orbit is highly circular and radial deviations of the model predictions are more easily visible.
With that said, a general case of energy conservation and overall shape agreement is still present.

\begin{figure}[htbp]
    \centering
    \includegraphics[width=\columnwidth]{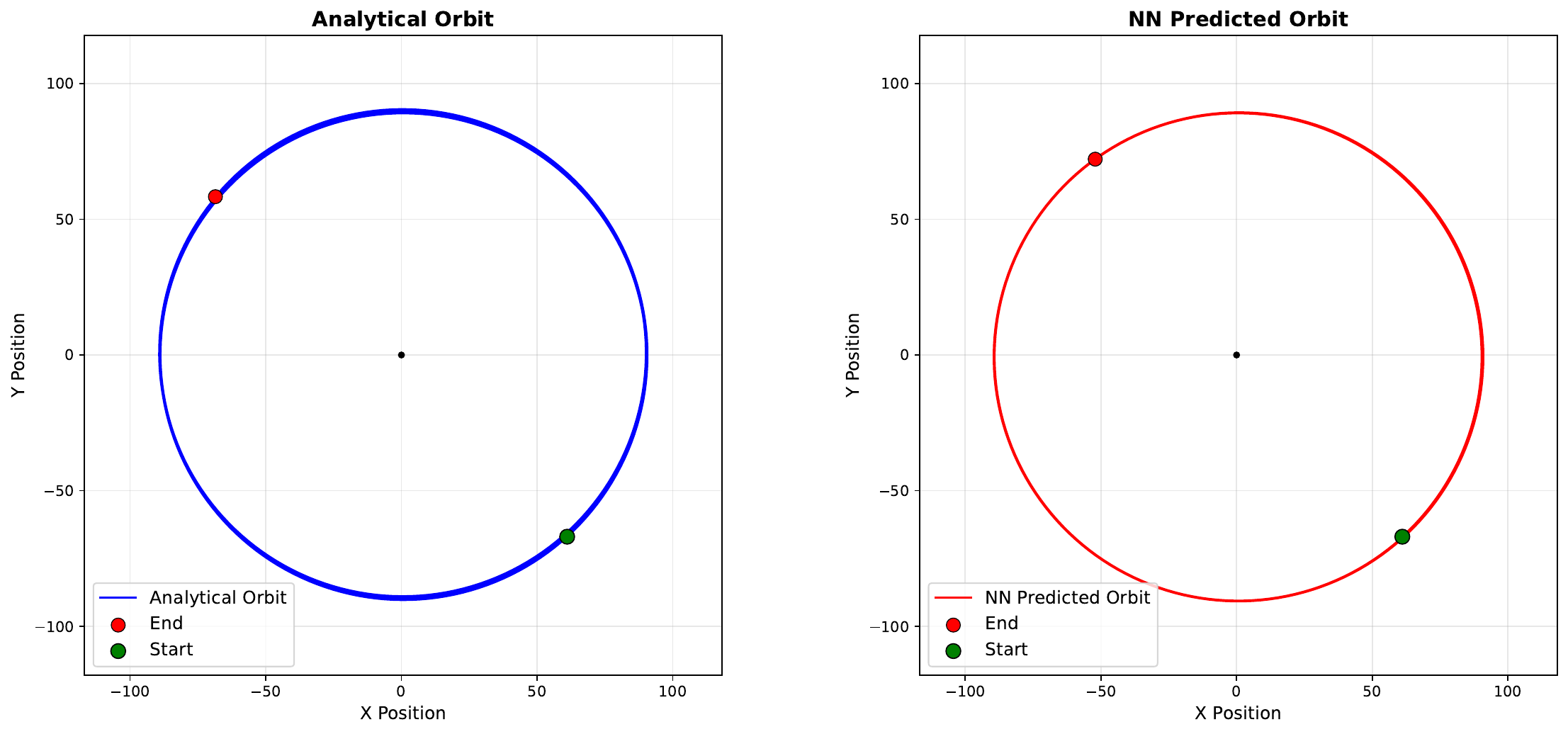}
    \caption{Side-by-side comparison for $e=\num{0.0101}$ test orbit in medium region.
Analytical to the left and model prediction to the right.
This orbit is highly circular which the model accurately predicts.}
    \label{fig:regular_circular_less_const3}
\end{figure}

%\begin{figure}[htbp]
 %   \centering
  %  \includegraphics[width=\columnwidth]{figures/regular_energy3.pdf}
   % \caption{Energy conservation error plot (top) and angular momentum conservation error (bottom) for medium region scenario 3.}
   % \label{fig:energy_cons_regular_less_const3}
%\end{figure}

\begin{figure}[htbp]
    \centering
    \includegraphics[width=\columnwidth]{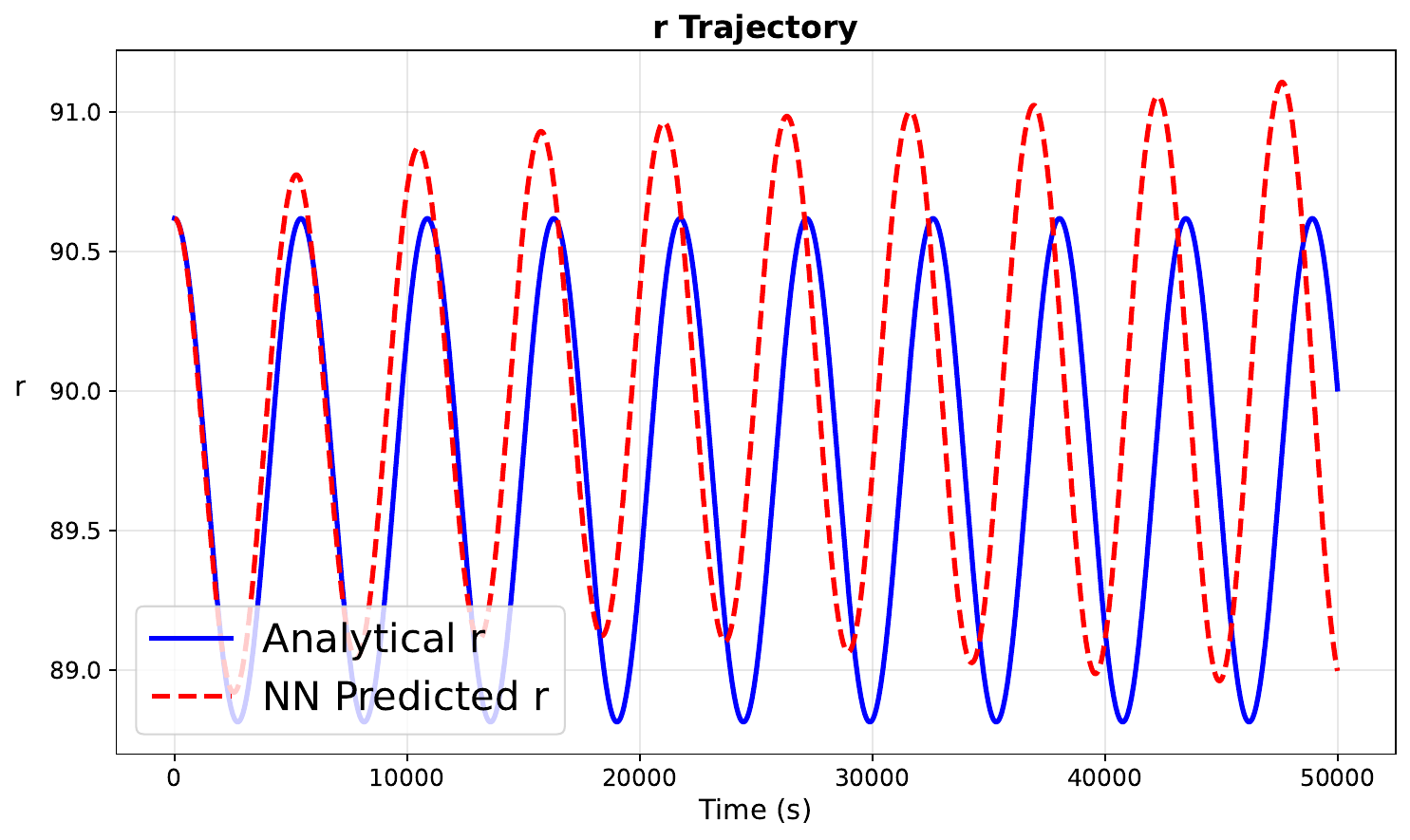}
    \caption{Evolution of the radial distance over time for analytical and predicted trajectory.
Accumulation of error is visible as the predicted orbit drifts and gains eccentricity.}
    \label{fig:r_circular_less_const3}
\end{figure}

\FloatBarrier
\subsubsection{Far Region:} 
Far region refers to the orbits with radial distance $r \in [\num{200}, \num{300}]$ for our training boundaries.
Relativistic effects of this region are much less dominant and can be considered the weak field limit of our Schwarzschild spacetime with $r_s=\num{2}$.

We tested two scenarios for this field one of which is a circular orbit test similar to the scenario 3 in medium region.
We present an example result of this scenario in the \cref{sec:predictions}.
Metrics for the scenarios and trajectory predictions for scenario 1 are given below.

\begin{table}[htbp]
\caption{Initial conditions and orbital metrics for far orbit scenarios.}
\label{tab:far_orbit_metrics}
\begin{ruledtabular}
\resizebox{\columnwidth}{!}{%    
\begin{tabular}{lc c c c c c}
 & \textbf{$\dot{r}$($\times 10^{-1}$)} & \textbf{$\dot{\phi}(\times 10^{-3})$} & \textbf{$L$} & \textbf{$E$} & \textbf{$e$} & \textbf{$r_{min}$ - $r_{max}$} \\
\colrule
1 & $[\num{0.1}, \num{0.2}]$ & $[\num{0.35}, \num{0.37}]$ & \num{14.728} & \num{0.9978} & \num{0.2347} & \numrange{173.19}{279.39} \\
2 & $[\num{0.0}, \num{0.01}]$ & $[\num{0.353}, \num{0.356}]$ & \num{14.232} & \num{0.9975} & \num{0.0082} & \numrange{197.91}{201.16} \\
\end{tabular}%
}
\end{ruledtabular}
\end{table}

\begin{figure}[htbp]
    \centering
    \includegraphics[width=\columnwidth]{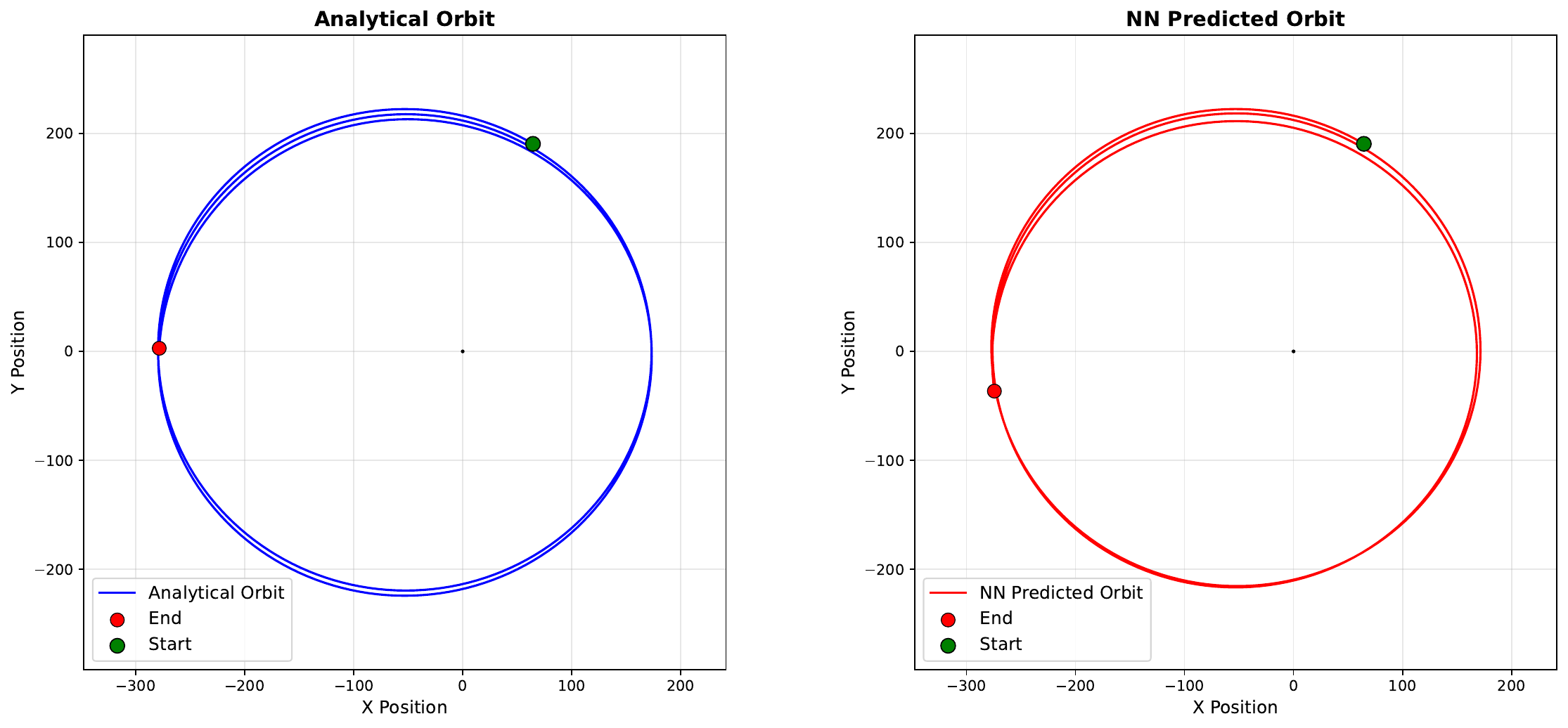}
    \caption{Side-by-side comparison for $e=\num{0.2347}$ test orbit.
Slight precession is visible in both analytical and predicted trajectories.}
    \label{fig:far_less_const3}
\end{figure}

Energy conservations are consistent with close and medium region experiments an we find it more useful to examine the spatial coordinate and velocity prediction accuracy of the model.
Specially the velocity metric gives hints of the weak yet present relativistic effects of the system.
\begin{figure}[htbp]
    \centering
    \includegraphics[width=\columnwidth]{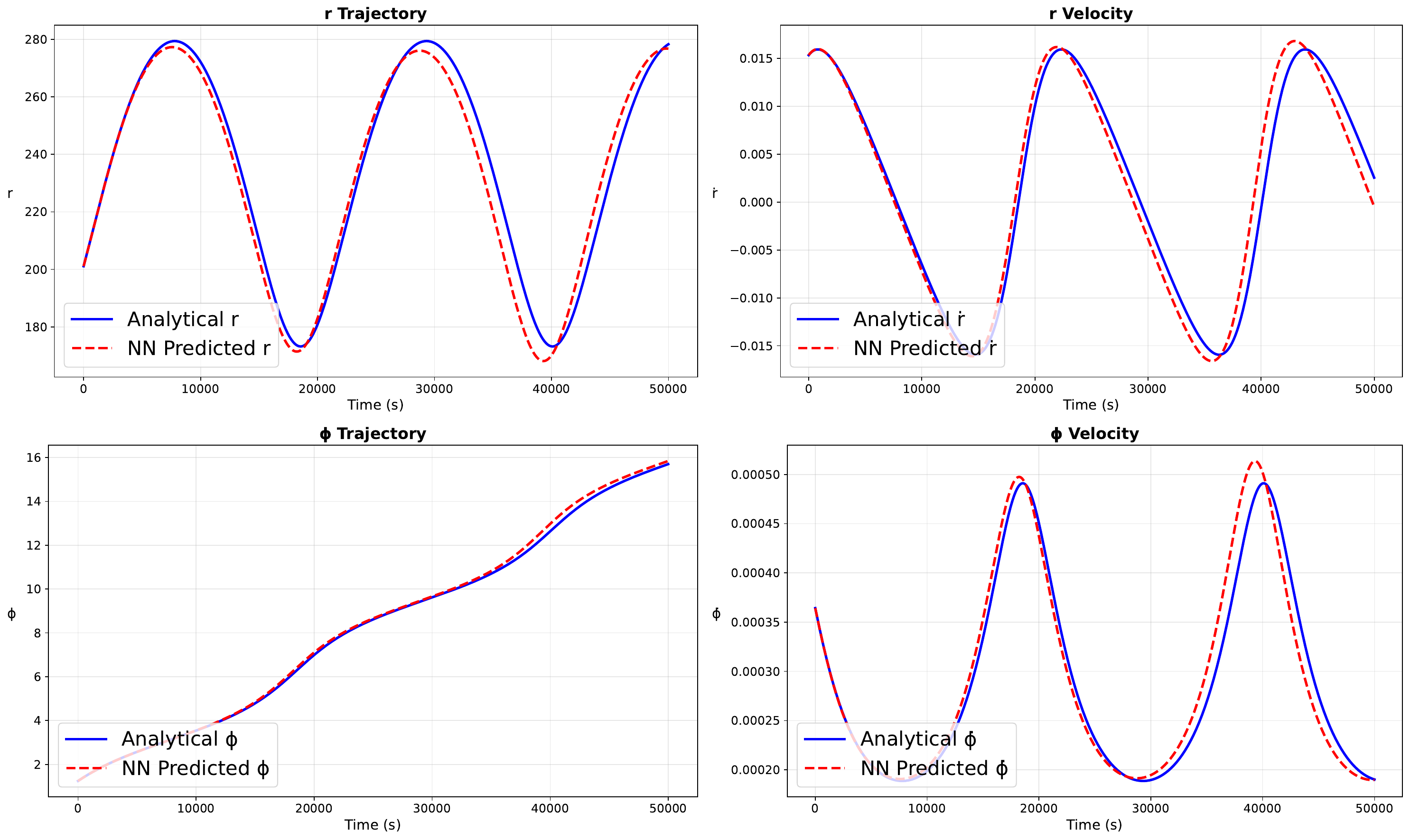}
    \caption{Spatial coordinate evolutions for analytical vs. predicted orbit with their velocities.}
    \label{fig:far_positions_less_const3}
\end{figure}

Scenarios presented here does not conclude the trajectory experiments and the rest can be found in the \cref{sec:predictions}.

\FloatBarrier
\begin{figure*}[htbp]
    \centering
    \includegraphics[width=\textwidth]{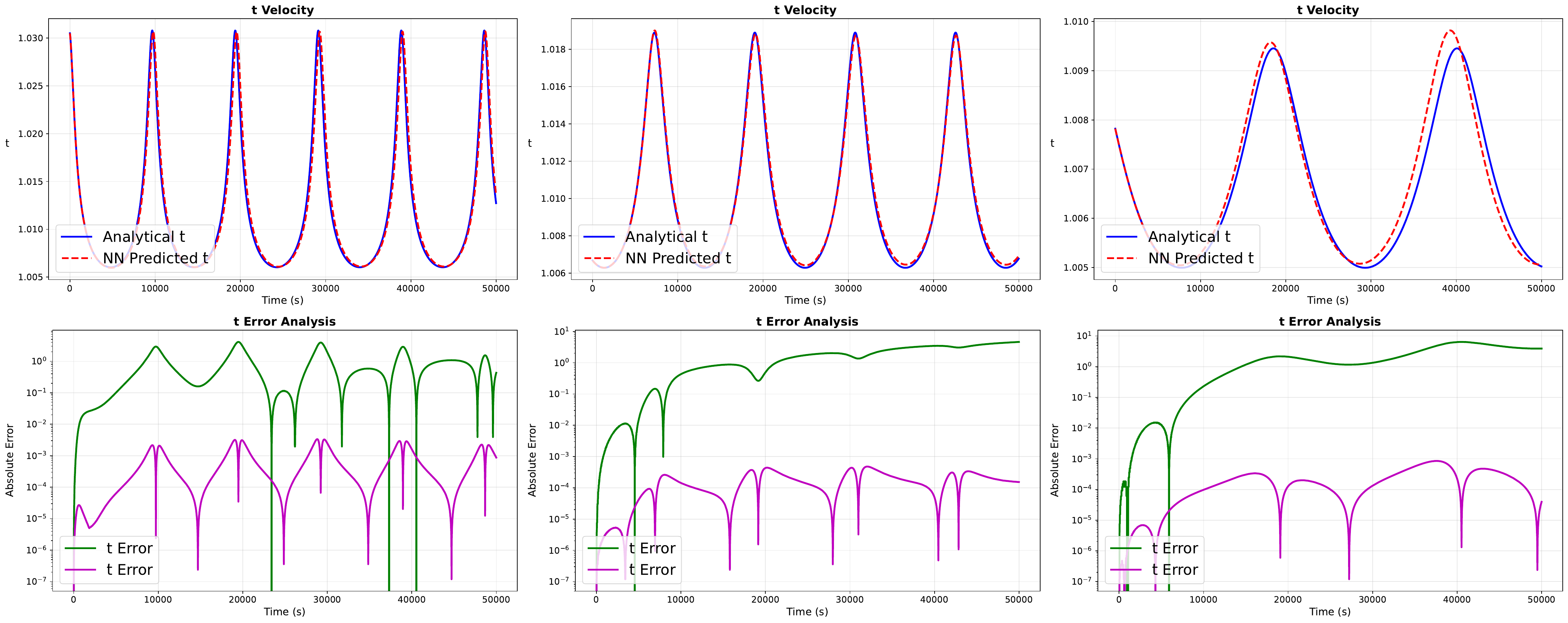}
    \caption{Time velocities for elliptical orbits from each region with their coordinate-time and time-velocity errors.
Shown here are: close region scenario 3 (left), medium region scenario 2 (middle) and far region scenario 1 (right).}
    \label{fig:t_vel}
\end{figure*}

\subsection{Time Dilation:}\label{sec:time_dilation}

Spatial dynamics of the geodesic motion is only half of the picture.
And one cannot talk about General Relativity without addressing the time dilation effects.
Time is also a dynamic coordinate of general-relativistic systems with its own speed and acceleration which our model respects equally as the spatial velocities and accelerations. Therefore we can examine the evolution of the rate of Schwarzschild coordinate time relative to the particle’s proper time as the time velocity $\dot{t}=\frac{dt}{d\tau}$. Here we present such evolutions for three trajectory scenarios, one for each region.

From \cref{eq:t_acell}, we read that time dilation effects get stronger as the particle gets closer to the primary body ($\ddot{t}$ increases)  and $\ddot{t}\propto\dot{r}$, therefore, the acceleration of this coordinate is most noticeable when the orbit is eccentric.
Thus, we focus on the evolution of time velocity for elliptical orbits all regions.

\begin{table}[htbp]
    \caption{Conserved specific energy, radial range, and expected range of
    $\dot{t}=dt/d\tau$. The maximum values are obtained from $\dot{t}=E/(1-r_s/r)$.}
    \label{tab:time_dilation_ranges}
    \begin{ruledtabular}
        \begin{tabular}{l c c c}
            \textbf{Scenario}
            & \textbf{$E$}
            & \textbf{Radial range}
            & \textbf{Max $\dot{t}$} \\
            \colrule
            Close   & $0.9969$ & $59.40$--$260.42$  & $1.0316$ \\
            Medium  & $0.9968$ & $94.83$--$218.39$  & $1.0183$ \\
            Far     & $0.9978$ & $173.19$--$279.39$ & $1.0095$ \\
        \end{tabular}
    \end{ruledtabular}
\end{table}

We find that the analytical and predicted curves agree closely in the close and medium regions. (see close region scenario 3 in \cref{fig:t_vel}).

In the far-region scenario, the model reproduces the overall periodic modulation but exhibits visible phase and amplitude deviations.
We attribute this discrepancy to limitations in the global loss optimization wich we discuss further in \cref{sec:extrapolation}.

\subsection{Physical Consistency Check and investigation of the acceleration as a function of input state:}\label{sec:accel_check}

\begin{figure}[htbp]
    \centering
    \includegraphics[width=\columnwidth]{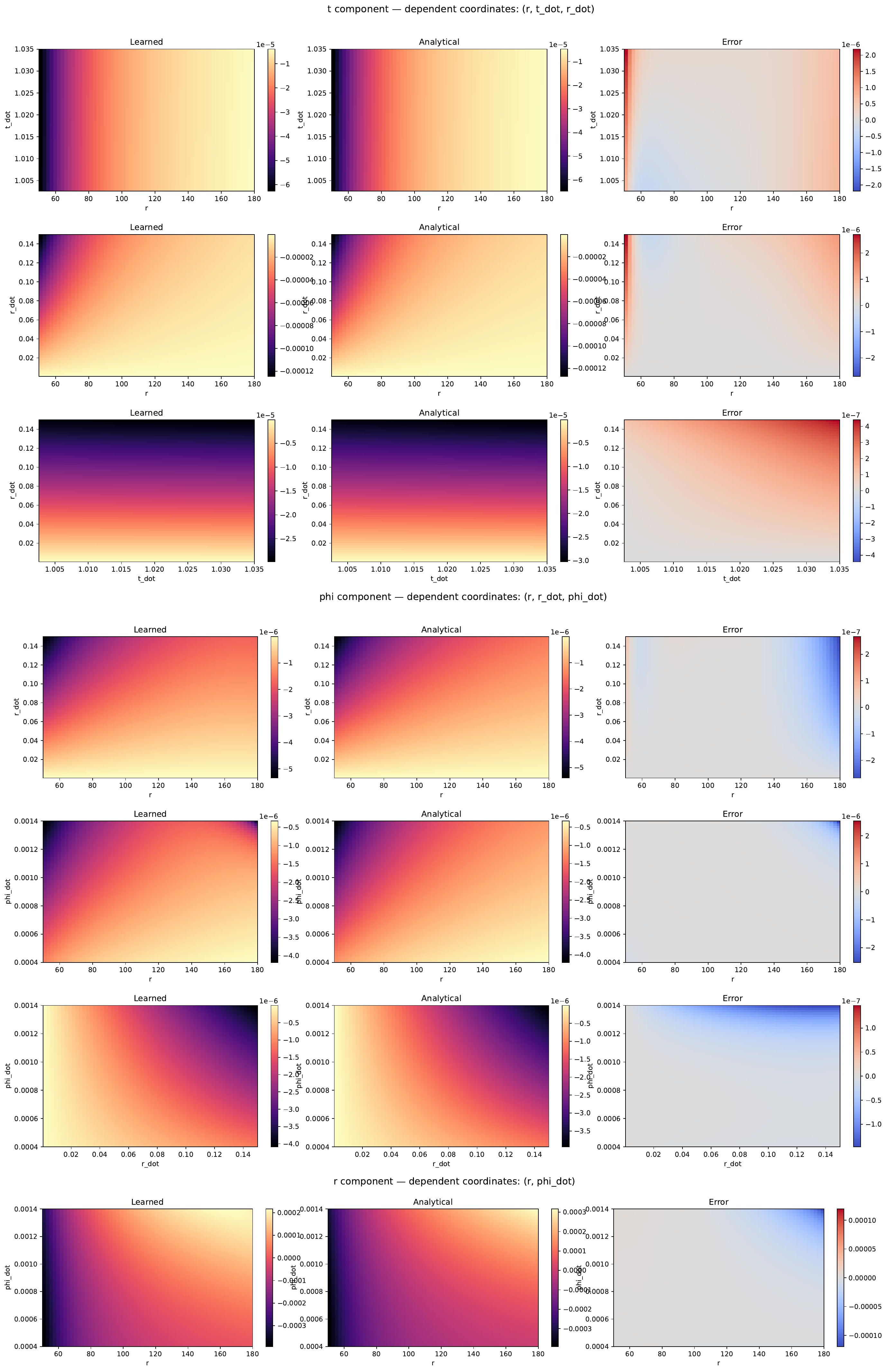}
    \caption{Acceleration components comparisons.
Plots for each component represents combinations of two coordinates on which the anaytical forms from \cref{eq:accel} is dependent.}
    \label{fig:accels}
\end{figure}

Trajectory scenarios only gives a limited access to the models overall picture about the dynamics. To see if the models acceleration predictions resemble the analytical form in \cref{eq:accel}, we can probe accelerations by looking at its behavior the training regime. 
Treating two of the input states as an independent variable $x,y$ and compare the shapes of the acceleration values function $f(x,y)$ reveals the functional agreement with the analytical form.
Here we present the coordinate dependence of each components of the acceleration for analytical functions and the models predictions along with their difference at \cref{fig:accels}.
To keep the simplicity, only the analytically relevant coordinates from \cref{eq:accel} are shown.

%\begin{figure*}[htbp]
 %   \centering
  %  \includegraphics[width=\textwidth]{figures/heatmap_fields_1x3_2.pdf}
   % \caption{Acceleration magnitudes in $r,\phi$ space for $r=[50,300]$ with fixed $\dot{r}$ and $\dot{\phi}$.}
    %\label{fig:acc_in_polar2}
%\end{figure*}
\subsection{Functional Analysis of Learned Lagrangian and Its Mass Matrix (Hessian):}
\label{sec:hessian_chapter}
Acceleration magnitudes comparison is not sufficient to claim the learned Lagrangian with which predicted functions for accelerations are derived using the Euler Lagrange equation is identical to the proposed solution \cref{eq:lagrangian}.
Statements about the Lagrangian forms themselves is not easy to make since the Euler-Lagrange equation is a second-order differential operator acting on $\mathcal{L}$; multiple Lagrangians can produce identical equations of motion~\cite{currie1966qequivalent,Lutter:2019ovq}.
Regardless, we find it informative to study the learned Lagrangian and its velocity Hessian, using the de-normalization pullback of Appendix~\ref{sec:norm} to compare learned and analytical quantities in a common physical coordinate system.

\subsubsection{Lagrangian:}\label{sec:lagrang}

We now probe the internal structure of the learned Lagrangian $L_{\rm learned}$ to determine whether it converges to the expected physical form of \cref{eq:lagrangian}

Lagrangian output from the model is a black box scalar function of input states.
Therefore, we can neither have any structure guesses nor any additional information other than a likely positive semi-definiteness due to the Sylvester regularization term described in \cref{sec:methods}.
It may also converge to \cref{eq:lagrangian} that has some useful restrictions like decoupled quadratic velocity terms that suggests a purely kinetic based Lagrangian, something we expect from a free particle following its geodesic.
However, this covnergence is not guarantied and the learned Lagrangian may not resemble the expected metric form \cref{eq:lagrangian} due to the inequivalence principle which suggests that the inverse problem of recovering $g_{\mu\nu}$ from $\mathcal{L}$ is not unique without further constraints \cite{carroll2019spacetime}.

\begin{figure}[htbp]
    \centering
    \includegraphics[width=\columnwidth]{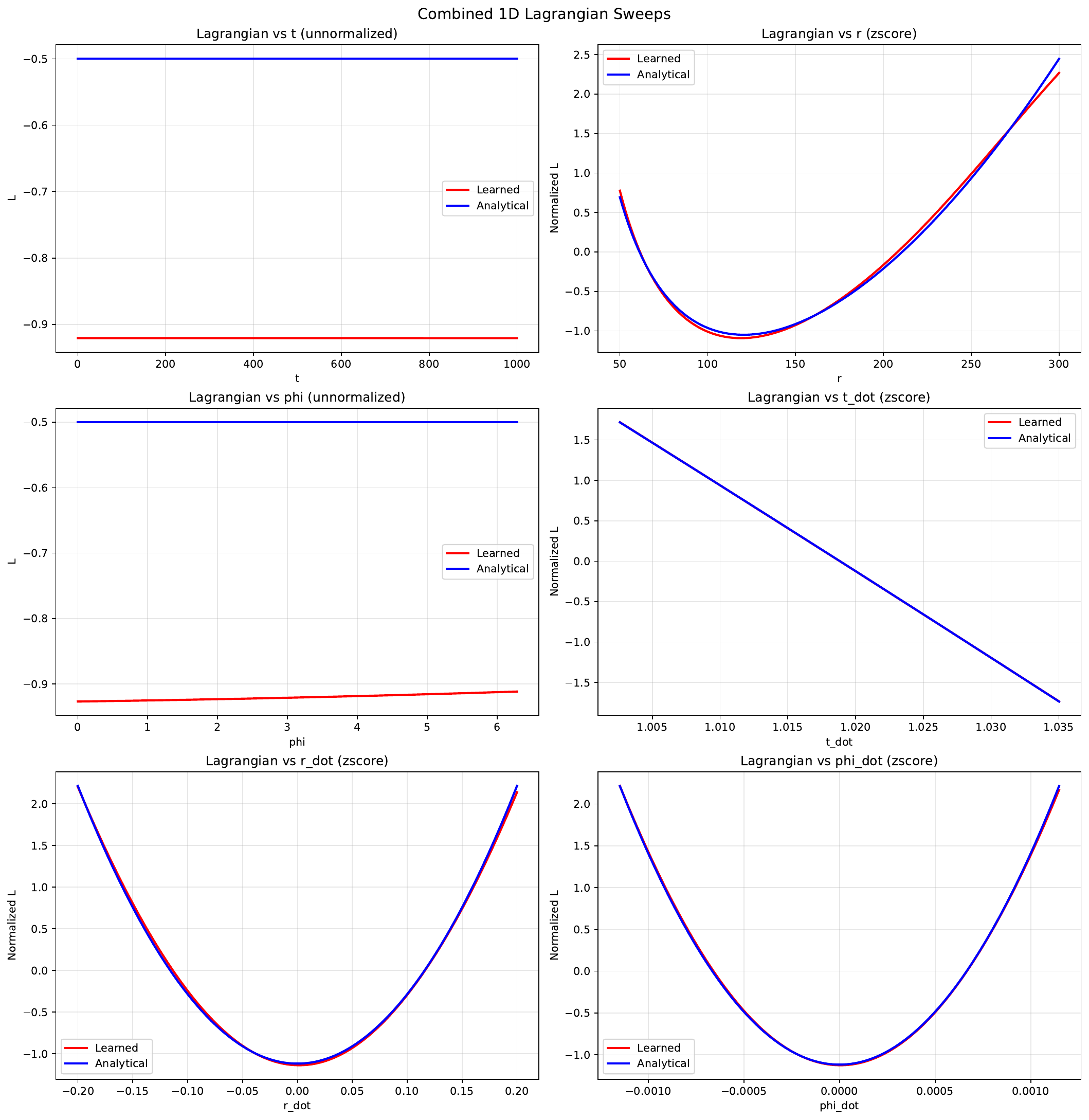}
    \caption{Lagrangian function comparisons in each coordinate.
t and $\phi$ plots are invariants of the system and the function should be constant under those coordinates.
Therefore, relevant plots has been kept unnormalized to see the deviations better.
The rest of the plots are normalized using the zscore for both analytical (blue) and learned functions (red), independently.}
    \label{fig:lag_1d}
\end{figure}

Therefore, the model is allowed to exploit the functional freedom to develop a different structure while still producing correct accelerations.
And similar to what we do with acceleration, we test this functional convergence by treating one of the variables inside the input state as an independent variable and compare the functional forms of analytical and predicted solutions under this dynamic coordinate.
Here we expect quadratic dependence in velocities due to the metric structure of \cref{eq:lagrangian} as well as constant behaviour for $t$ and $\phi$ due to the system's time and rotation symmetries.
In \cref{fig:lag_1d}, we see such comparison where each of the 6 input state coordinate is treated as the dynamic variable.

\begin{figure}[htbp]
    \centering
    \includegraphics[width=\columnwidth]{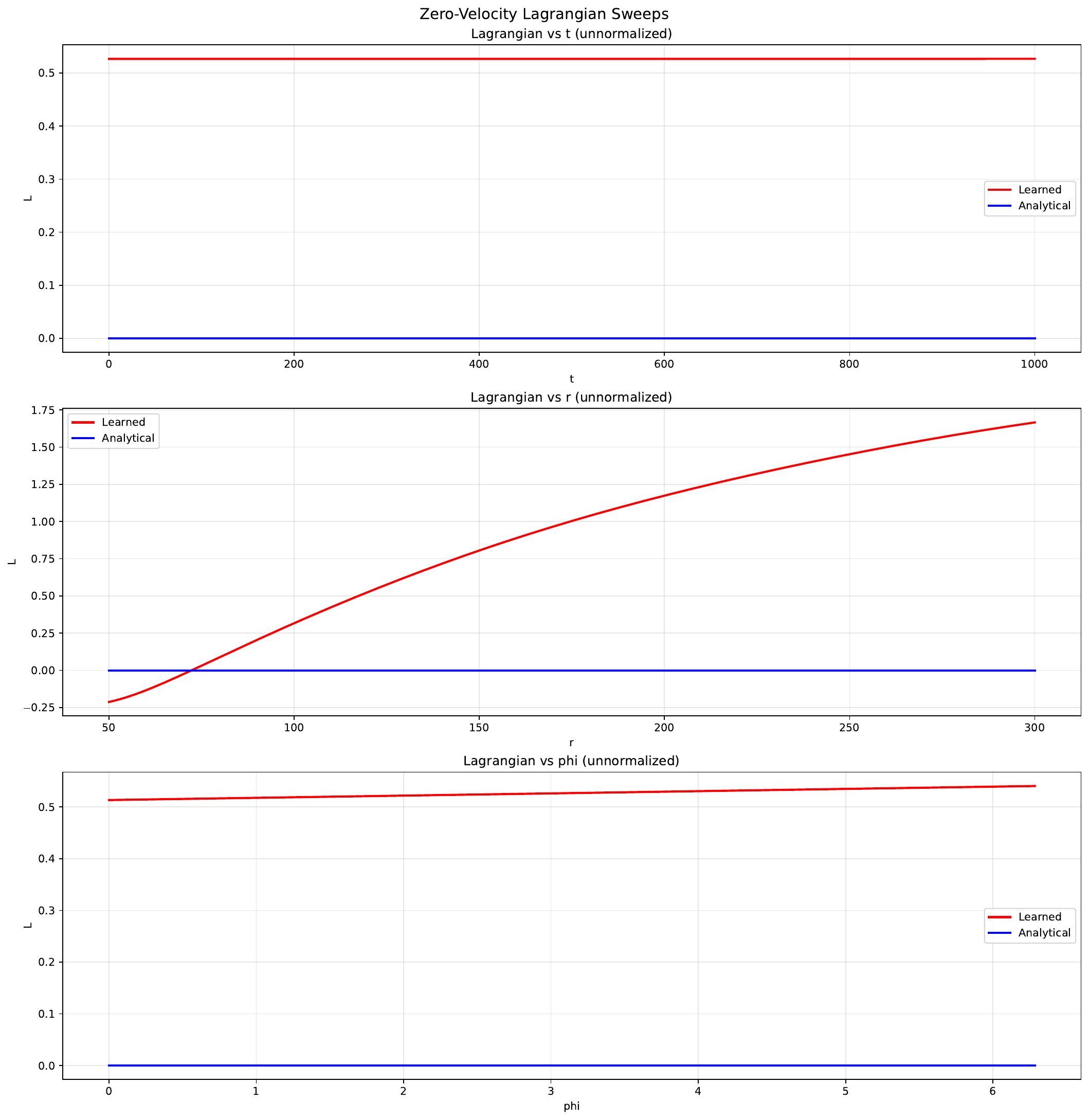}
    \caption{0 velocity Lagrangian plots.
Learned function contains r and phi dependent additional terms that is not present in the analytical form.}
    \label{fig:lag_zero}
\end{figure}

\begin{figure*}[htbp]
    \centering
    \includegraphics[width=\textwidth]{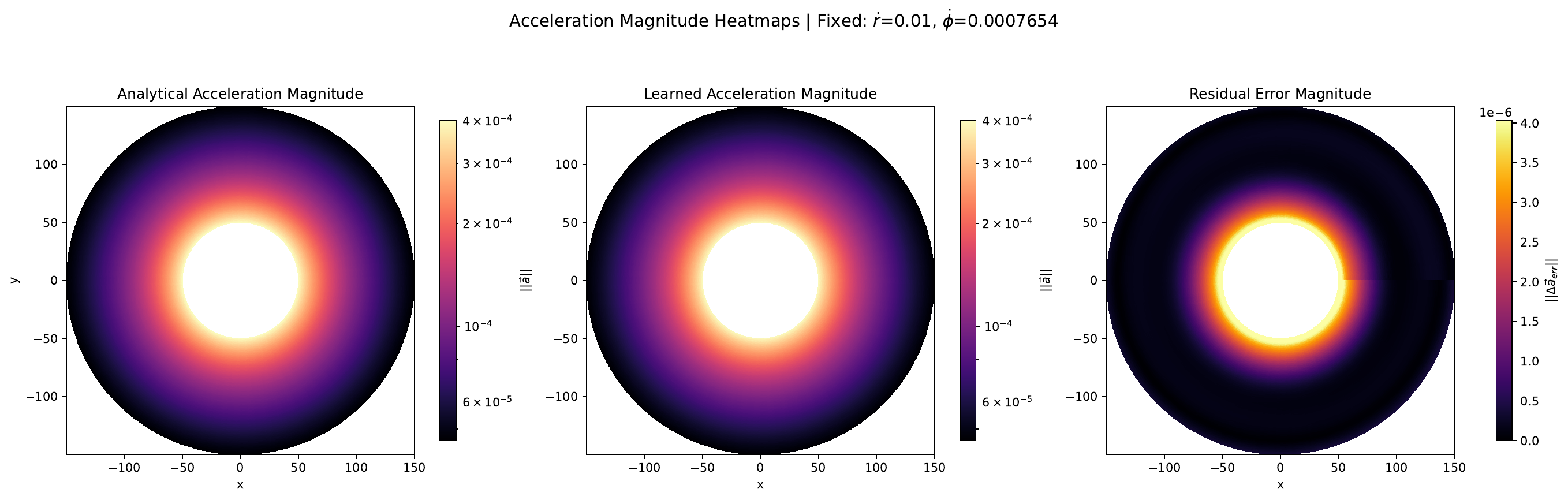}
    \caption{Acceleration magnitudes in $r,\phi$ space for $r=[50,150]$ with fixed $\dot{r}$ and $\dot{\phi}$.}
    \label{fig:acc_in_polar}
\end{figure*}

From \cref{fig:acc_in_polar,fig:lag_zero}, we notice a slight phi dependence in the learned system, that hides itself when the model is used to predict trajectories.
Deviations are small to have an impact on the predictions yet still there.
Too see this better, we also looked at a scenario where $\dot{t}=\dot{r}=\dot{\phi}=0$.
Under this condition, we expect a constant function from the model if it converged to a truly kinetic based Lagrangian that does not depend on a "potential-like" additional position terms.
We observe two deviations in $\phi$ and $r$, the latter being much significant.

The evidence from the one-dimensional Lagrangian slices (\cref{fig:lag_1d} and \cref{fig:lag_zero}) is consistent with a learned Lagrangian of the approximate form
\begin{align}
\label{eq:learned_lag}
    \mathcal{L}_{learned}=&-A(r,\phi,\dot{t})\dot{t}^2+B(r,\phi,\dot{t})\dot{r}^2+C(r,\phi,\dot{t})\dot{\phi}^2 \nonumber \\
    &+V(r,\phi)
\end{align}
However, full functional characterization would require probing the full 6D input space, which is outside the scope of this work.

\subsubsection{Hessian:}
The mass matrix (velocity Hessian) of the Lagrangian is defined as:
    \begin{equation}
        H_{ij} = \frac{\partial^2 \mathcal{L}}{\partial \dot{q}^i \partial \dot{q}^j}.
    \end{equation}
    For a free particle on a geodesic, the Lagrangian takes the purely kinetic, metric-based form $\mathcal{L} = \frac{1}{2}g_{\alpha\beta}(x)\dot{x}^\alpha \dot{x}^\beta$ [Eq.~\eqref{eq:lag}]. Differentiating with respect to the velocities explicitly recovers the spacetime metric tensor:
    \begin{equation}
        \label{eq:hess}
        H_{\mu\nu} = \frac{\partial^2}{\partial \dot{x}^\mu \partial \dot{x}^\nu} \left( \frac{1}{2} g_{\alpha\beta}(x)\dot{x}^\alpha \dot{x}^\beta \right) = g_{\mu\nu}(x).
    \end{equation}
    Equivalently, in coordinate-free matrix notation with $\mathcal{L} = \frac{1}{2}\dot{\mathbf{q}}^\top \mathbf{g}(\mathbf{q})\dot{\mathbf{q}}$:
    \begin{equation}
        \mathbf{H} = \nabla_{\dot{\mathbf{q}}}\nabla_{\dot{\mathbf{q}}}^\top \mathcal{L} = \mathbf{g}(\mathbf{q}).
    \end{equation}

Applying $n$ dimensional velocity gradient operator twice to our output Lagrangian function, we obtain $n^2$ number of Hessian functions, all of which shares the same input state as the $\mathcal{L}_{learned}$.
We like to highlight the derivative scaling due to normalization needs to be addressed if we are to compare this function with a function that uses derivatives in unscaled real-world coordinates.
The exact scales to de-normalize our learned $H$ functions is outlined in \cref{sec:norm}.

The analytical hessian is connected to the metric of space $g$ from \cref{eq:hess}.
Therefore, it is only a function of r like described in \cref{eq:metric}.
Using this knowledge, we used a similar trick to the previous section to show the relation for the de-normalized learned hessian function with respect to r to compare with the analytical form.
For a comprehensive analysis, all the forms as different functions of state variables should be investigated.

\begin{figure}[htbp]
 %   \centering
  %  \includegraphics[width=\columnwidth]{figures/hess_raw_comparison.pdf}
   % \caption{Hessian comparisons in real magnitudes.
%off-diagonal non zero components are present.}
 %   \label{fig:hess_raw}

    \centering
    \includegraphics[width=\columnwidth]{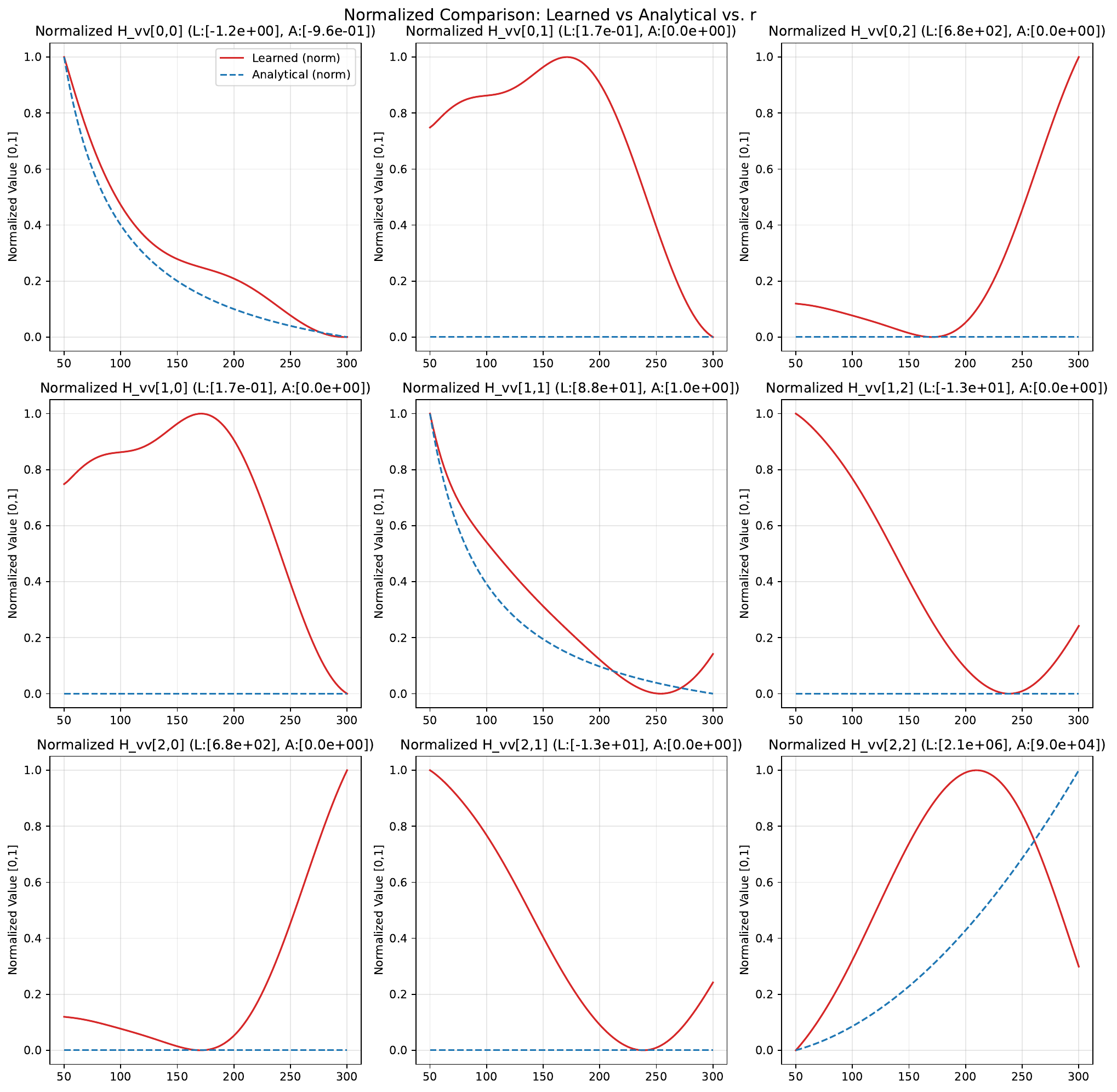}
    \caption{Normalized Hessian components as function of r.
Diagonal mismatch is most present in $\phi,\phi$ component.}
    \label{fig:hess_norm}
\end{figure}

However, we find it sufficient to show an r-form discrepancy to claim the functional form of the Lagrangian does not inherit the properties of the purely kinetic form \cref{eq:lag}.
From \cref{fig:hess_norm}, we see non-zero cross term magnitudes in $H_{\dot{t},\dot{\phi}}$ and $H_{\dot{r},\dot{\phi}}$ which is in line with the predictions from \cref{eq:learned_lag}.In that prediction, quadratic relation in $\dot{t}$ could not be confirmed due to small testing region in that coordinate.
Now the conclusion about the coefficients $A,B$ and $C$ is more justified with non-zero derivatives in the cross terms.
\FloatBarrier

\section{Summary of Findings and Conclusions}\label{sec:conclusion}

\subsection{Summary of Contributions:}

We have presented and evaluated two improvements to the stabilized LNN framework for Schwarzschild geodesic learning: 
\begin{itemize}
    \item a  decoupled position-velocity normalization scheme
    \item an energy-constrained velocity sampling strategy
\end{itemize}
Together, these address the core failure modes identified in the baseline approach.
The decoupled normalization resolves the numerical imbalance between coordinates spanning disparate scales and enabling convergence when the dynamics of the system spans multiple regimes.
The energy-constrained sampling, particularly the mixed velocities strategy that combines two radial velocity windows, provides the most uniform coverage of the velocity-phase space and raises the fraction satisfying $E<1$ from \qty{27.3}{\percent} to \qty{92.3}{\percent}, improving trajectory prediction fidelity in bounded test orbits.

\subsection{Trajectory Predictions:}

Across three orbital regimes and eight scenarios close ($r \in [50,90]$), medium ($r \in [90,200]$), and far ($r \in [200,300]$) our best model reproduces orbital paths with correct precession (\cref{fig:traj_accuracy}) with average radial deviation of $\num{4.7662}  \pm  \num{4.4051}$ in $r_s$ units, maintains energy and angular momentum conservation to within a few percent over $5\times10^4$s of integration, and accurately captures time dilation effects.

Notably, we observe a dissociation between training loss and trajectory quality (\cref{fig:losses_all}). The baseline dataset achieves faster apparent loss decay by exploiting high-magnitude unbound samples, while the mixed velocity dataset forces the model to fit the harder low-velocity bound regimes that are critical for stable integration.
This highlights a tradeoff between training dataset quality and loss minimization that translates directly to trajectory fidelity(\cref{fig:norm_effects}). 

The general conclusion for the trajectory experiments is towards the models success of learning the spacial part of the orbital dynamics.
We also raise attention to the effects of error accumulation which presents itself for highly circular or far orbits.
As the orbital dynamics speed up in the close region, predictions become more stable and dynamics get more accurate.

\section{Future Work and Limitations}\label{sec:future_work}
\subsection{Physical Interpretability:}\label{sec:extrapolation}
While our model accurately reproduces Schwarzschild geodesics within the training radial range $r \in [50, 300]$ (in units of $r_s/2$), the physical interpretability of the learned dynamics hinges on whether a more generalized model can successfully learn dynamics of both strong and weak field regimes where observational data being more abundant in the latter. Extrapolating to length scales far outside the training distribution raises the question of how faithfully the learned effective Lagrangian captures the underlying spacetime structure in the weak-field limit.

For comparison, in the geometric units adopted here ($r_s=2$), Mercury's semi-major axis corresponds to $r \approx \num{4e7}$. Five orders of magnitude beyond our training range shows the scale separation between the strong-field dynamics accessible to the model and the weak-field regime relevant to observational solar-system scenarios. 

We already observe some of the difficulties of multi-regional learning task in our training boundaries. Looking at \cref{fig:t_vel}, we see that the time coordinate velocity predictions degrade slightly when we get further from the primary body. We attribute this to difficulties in global loss optimization where the infinitesimal variations in weak-field time acceleration producing gradients too small to effectively signal the network as they 
are overshadowed by the more dominant error signals from the strong-field regime, which is a common problem in multi-regional learning tasks. \cite{kendall2018multi,chen2018gradnorm}
For further outer regions of the system, the time velocity starts from $\approx \num{1.005}$ and approaches to $E$ (refer to \cref{eq:killing1}) as the spacetime becomes asymptotically flat. Notice $\dot{t}=1$ implies one-to-one correspondence with the affine proper time parameter $\tau$ which an unbound particle at infinity would experience. For comparison, this ratio is $\approx \num{1.030}$ when $r\approx\num{50}$. 

Since weak field regime of Schwarzschild dynamics is mostly Newtonian-like and does not show strong general relativistic effects, we expect training of a similar network on this regime to be a more straightforward task compared to what this work presents. However, a model capable of both capturing the strong field dynamics and be able to accurately represent systems with realistic observational ranges is more involved and has limitations in training as we discussed. As a result, the long-term goal of applying LNN-based methods to real astronomical ranges and observations while correctly representing the strong relativistic effects remains an important direction for future work.

\subsection{Inequivalent Lagrangians}

One might hope that the learned Lagrangian converges to a nondegenerate, quadratic in the velocities form:
\begin{equation}
    \mathcal{L}_h
    = \frac{1}{2}h_{\mu\nu}(x)\dot{x}^{\mu}\dot{x}^{\nu},
\end{equation}
for which the Euler-Lagrange equations are the affinely parameterized
geodesic equations associated with $h_{\mu\nu}$. Reproducing the
Schwarzschild affine connection in our studies case therefore requires
$\nabla^{(S)}_{\alpha}h_{\mu\nu}=0$. Since Schwarzschild spacetime
has full Lorentz holonomy (type $R_{15}$), the only compatible
metric is a constant nonzero multiple of the Schwarzschild metric,
$h_{\mu\nu}=c\,g^{(S)}_{\mu\nu}$ \cite{hall1993spacetimes}.
This uniqueness applies only within the class of nondegenerate
velocity-quadratic Lagrangians and for such Lagrangians, the velocity Hessian must be proportional to the metric $g^{(S)}_{\mu\nu}$.

However, convergence to this exact physical form is not guaranteed, as the model is free to deviate during training toward any mathematically equivalent Lagrangian. 
Moreover, the decoupled normalization leads the model to learn a latent scalar $\mathcal{L}_\theta$ in normalized coordinates $(\tilde{q},\tilde{\dot{q}})$. 
As argued in Sec.~\ref{sec:normal} and Appendix~\ref{sec:norm}, the model is free to develop any functional form that produces $\tilde{\ddot{q}}_{\rm true}$. 

Based on the investigation in \cref{sec:lagrang}, we conclude that the learned Lagrangian does not converge to the purely kinetic form. The velocity Hessian develops non-zero off-diagonal components, and the Lagrangian acquires a residual position-dependent term $V(r,\phi)$ visible even at zero velocity.

Consequently, while our method reliably predicts Schwarzschild geodesic dynamics, extracting the spacetime metric directly from the learned Lagrangian requires additional structural constraints beyond those imposed here. 

This metric extraction task, although the metric being not inherently guaranteed to be the one that correctly represents the geometry of the system due to the blackbox aproach of LNN's, remains a worthwhile endeavor. And converging to a desired form of Lagrangian whether by further conditioning the Lagrangian model or adding regularizations could make it possible to extract interpretable metric elements, which remains as a future work.

\bibliographystyle{apsrev4-2}
\bibliography{references}

@article{Schwarzschild1916,
  title = {\"Uber Das {{Gravitationsfeld}} Eines {{Massenpunktes}} Nach Der {{Einsteinschen Theorie}}},
  author = {Schwarzschild, Karl},
  year = 1916,
  month = jan,
  journal = {Sitzungsberichte der K\"oniglich Preussischen Akademie der Wissenschaften},
  pages = {189--196},
  urldate = {2025-11-30}
}

@article{currie1966qequivalent,
  title = {q-{{Equivalent Particle Hamiltonians}}. {{I}}. {{The Classical One-Dimensional Case}}},
  author = {Currie, D. G. and Saletan, E. J.},
  year = 1966,
  journal = {J. Math. Phys.},
  volume = {7},
  number = {6},
  pages = {967--974},
  doi = {10.1063/1.1705009},
}

@inproceedings{Lutter:2019ovq,
  title     = {Deep Lagrangian Networks:
               Using Physics as Model Prior for Deep Learning},
  author    = {Lutter, Michael and Ritter, Christian and Peters, Jan},
  booktitle = {International Conference on Learning Representations},
  year      = {2019},
  eprint    = {1907.04490},
  archiveprefix = {arXiv},
  primaryclass  = {cs.LG},
  url       = {https://openreview.net/forum?id=BklHpjCqKm}
}

@misc{Hamzaogullari:2026,
  title  = {Learning Relativistic Geodesics and Chaotic Dynamics
            via Stabilized Lagrangian Neural Networks},
  author = {Hamzaogullari, Abdullah Umut and Ozakin, Arkadas},
  year   = {2026},
  month  = jan,
  eprint = {2601.12519},
  archiveprefix = {arXiv},
  primaryclass  = {cs.LG},
  doi = {10.48550/arXiv.2601.12519}
}

@article{Lim:2024mkb,
  title = {Energies and Angular Momenta of Periodic {{Schwarzschild}} Geodesics},
  author = {Lim, Yen-Kheng and Yeo, Zhi Cheng},
  year = 2024,
  month = jan,
  journal = {Phys. Rev. D},
  volume = {109},
  number = {2},
  eprint = {2401.13894},
  primaryclass = {gr-qc},
  pages = {024037},
  issn = {2470-0010, 2470-0029},
  doi = {10.1103/PhysRevD.109.024037},
  urldate = {2026-03-12},
  archiveprefix = {arXiv}
}

@article{raissiPhysicsInformedNeural2019,
  title   = {Physics-Informed Neural Networks:
             A Deep Learning Framework for Solving Forward and Inverse
             Problems Involving Nonlinear Partial Differential Equations},
  author  = {Raissi, Maziar and Perdikaris, Paris and
             Karniadakis, George E.},
  journal = {J. Comput. Phys.},
  volume  = {378},
  pages   = {686--707},
  year    = {2019},
  doi     = {10.1016/j.jcp.2018.10.045}
}

@article{lunaSolvingTeukolsky2023,
  title   = {Solving the {Teukolsky} Equation with
             Physics-Informed Neural Networks},
  author  = {Luna, Raimon and Calder{\'o}n Bustillo, Juan and
             Seoane Mart{\'i}nez, Juan Jos{\'e} and
             Torres-Forn{\'e}, Alejandro and Font, Jos{\'e} A.},
  journal = {Phys. Rev. D},
  volume  = {107},
  number  = {6},
  pages   = {064025},
  year    = {2023},
  doi     = {10.1103/PhysRevD.107.064025},
  eprint  = {2212.06103},
  archiveprefix = {arXiv},
  primaryclass  = {gr-qc}
}

@misc{liSolvingEinstein2023,
  title  = {Solving {Einstein} Equations Using Deep Learning},
  author = {Li, Zhi-Han and Li, Chen-Qi and Pang, Long-Gang},
  year   = {2023},
  eprint = {2309.07397},
  archiveprefix = {arXiv},
  primaryclass  = {gr-qc},
  doi = {10.48550/arXiv.2309.07397}
}

@misc{schwarzschild1999gravitational,
  title = {On the Gravitational Field of a Mass Point According to {{Einstein}}'s Theory},
  author = {Schwarzschild, K.},
  year = 1999,
  month = may,
  number = {arXiv:physics/9905030},
  eprint = {physics/9905030},
  publisher = {arXiv},
  doi = {10.48550/arXiv.physics/9905030},
  urldate = {2026-03-13},
  archiveprefix = {arXiv}
}

@inproceedings{chenNeuralOrdinaryDifferential2019,
  title     = {Neural Ordinary Differential Equations},
  author    = {Chen, Ricky T. Q. and Rubanova, Yulia and
               Bettencourt, Jesse and Duvenaud, David K.},
  booktitle = {Advances in Neural Information Processing Systems},
  volume    = {31},
  year      = {2018},
  publisher = {Curran Associates, Inc.},
  eprint    = {1806.07366},
  archiveprefix = {arXiv},
  primaryclass  = {cs.LG}
}

@misc{cranmerDiscoveringSymbolicModels2020,
  title = {Discovering {{Symbolic Models}} from {{Deep Learning}} with {{Inductive Biases}}},
  author = {Cranmer, Miles and {Sanchez-Gonzalez}, Alvaro and Battaglia, Peter and Xu, Rui and Cranmer, Kyle and Spergel, David and Ho, Shirley},
  year = 2020,
  month = nov,
  number = {arXiv:2006.11287},
  eprint = {2006.11287},
  primaryclass = {cs},
  publisher = {arXiv},
  doi = {10.48550/arXiv.2006.11287},
  urldate = {2026-03-30},
  archiveprefix = {arXiv}
}

@misc{cranmerLagrangianNeuralNetworks2020,
  title = {Lagrangian {{Neural Networks}}},
  author = {Cranmer, Miles and Greydanus, Sam and Hoyer, Stephan and Battaglia, Peter and Spergel, David and Ho, Shirley},
  year = 2020,
  month = jul,
  number = {arXiv:2003.04630},
  eprint = {2003.04630},
  primaryclass = {cs},
  publisher = {arXiv},
  doi = {10.48550/arXiv.2003.04630},
  urldate = {2025-08-29},
  archiveprefix = {arXiv}
}

@misc{finziSimplifyingHamiltonianLagrangian2020,
  title = {Simplifying {{Hamiltonian}} and {{Lagrangian Neural Networks}} via {{Explicit Constraints}}},
  author = {Finzi, Marc and Wang, Ke Alexander and Wilson, Andrew Gordon},
  year = 2020,
  month = oct,
  number = {arXiv:2010.13581},
  eprint = {2010.13581},
  primaryclass = {cs},
  publisher = {arXiv},
  doi = {10.48550/arXiv.2010.13581},
  urldate = {2026-03-30},
  archiveprefix = {arXiv}
}

@inproceedings{greydanusHamiltonianNeuralNetworks2019,
  title     = {Hamiltonian Neural Networks},
  author    = {Greydanus, Samuel and Dzamba, Misko and Yosinski, Jason},
  booktitle = {Advances in Neural Information Processing Systems},
  volume    = {32},
  pages     = {15353--15363},
  year      = {2019},
  publisher = {Curran Associates, Inc.},
  eprint    = {1906.01563},
  archiveprefix = {arXiv},
  primaryclass  = {cs.LG}
}

@inproceedings{hall1993spacetimes,
    author    = {Hall, G. S.},
    title     = {Space-times and Holonomy Groups},
    booktitle = {Differential Geometry and Its Applications:
                 Proceedings of the Fifth International Conference,
                 Opava, August 24--28, 1992},
    editor    = {Kowalski, O. and Krupka, D.},
    pages     = {201--210},
    publisher = {Silesian University},
    address   = {Opava},
    year      = {1993}
}

@inproceedings{chen2018gradnorm,
  title     = {{GradNorm}: Gradient Normalization for Adaptive Loss Balancing in Deep Multitask Networks},
  author    = {Chen, Zhao and Badrinarayanan, Vijay and Lee, Chen-Yu and Rabinovich, Andrew},
  booktitle = {Proceedings of the 35th International Conference on Machine Learning (ICML)},
  year      = {2018},
  pages     = {794--803},
  url       = {https://proceedings.mlr.press/v80/chen18a.html}
}

@inproceedings{kendall2018multi,
  title     = {Multi-Task Learning Using Uncertainty to Weigh Losses for Scene Geometry and Semantics},
  author    = {Kendall, Alex and Gal, Yarin and Cipolla, Roberto},
  booktitle = {Proceedings of the IEEE Conference on Computer Vision and Pattern Recognition (CVPR)},
  year      = {2018},
  pages     = {7482--7491},
  doi       = {10.1109/CVPR.2018.00781}
}

@book{carroll2019spacetime,
  title     = {Spacetime and Geometry: An Introduction to General Relativity},
  author    = {Carroll, Sean M.},
  year      = {2019},
  publisher = {Cambridge University Press},
  isbn      = {978-1-108-48839-6}
}

@misc{gherardiniBlackHoleBlack2026,
  title  = {Black Hole Black Boxes:
            Numerical Black Hole Metrics via {AInstein} Neural Networks},
  author = {Schettini Gherardini, Tancredi and Hirst, Edward and
            Stapleton, Alexander George},
  year   = {2026},
  month  = jul,
  eprint = {2607.05489},
  archiveprefix = {arXiv},
  primaryclass  = {gr-qc},
  doi = {10.48550/arXiv.2607.05489}
}

@incollection{lecun1998efficient,
    author    = {LeCun, Yann and Bottou, L{\'e}on and Orr, Genevieve B.
                 and M{\"u}ller, Klaus-Robert},
    title     = {Efficient BackProp},
    booktitle = {Neural Networks: Tricks of the Trade},
    editor    = {Orr, Genevieve B. and M{\"u}ller, Klaus-Robert},
    series    = {Lecture Notes in Computer Science},
    volume    = {1524},
    pages     = {9--50},
    publisher = {Springer},
    address   = {Berlin, Heidelberg},
    year      = {1998},
    doi       = {10.1007/3-540-49430-8_2}
}

@article{sola1997importance,
    author  = {Sola, J. and Sevilla, J.},
    title   = {Importance of Input Data Normalization for the Application
               of Neural Networks to Complex Industrial Problems},
    journal = {IEEE Transactions on Nuclear Science},
    volume  = {44},
    number  = {3},
    pages   = {1464--1468},
    year    = {1997},
    doi     = {10.1109/23.589532}
}

@article{do2019redshift,
  title   = {Relativistic redshift of the star S0-2 orbiting the
             Galactic center supermassive black hole},
  author  = {Do, Tuan and others},
  journal = {Science},
  volume  = {365},
  number  = {6454},
  pages   = {664--668},
  year    = {2019},
  doi     = {10.1126/science.aav8137}
}

@article{gravity2020precession,
  author  = {{GRAVITY Collaboration}},
  title   = {Detection of the Schwarzschild precession in the orbit
             of the star S2 near the Galactic centre massive black hole},
  journal = {Astronomy \& Astrophysics},
  volume  = {636},
  pages   = {L5},
  year    = {2020},
  doi     = {10.1051/0004-6361/202037813}
}

\appendix
\crefalias{section}{appendix}
\crefalias{section}{appendix}
% Reset float counters with an "A" prefix for appendix:
\renewcommand{\thefigure}{A\arabic{figure}}
\setcounter{figure}{0}
\renewcommand{\thetable}{A\Roman{table}}
\setcounter{table}{0}

\section{Setup for the Model Training, Dataset Variances and performance comparisons}
\label{sec:setup}

\subsection{Model Architecture:}

Table~\ref{tab:architecture} summarizes the architecture and training hyperparameters used in all experiments.

\begin{table}[htbp]
\caption{Model architecture and training hyperparameters. The values shown are those used to produce the main results.}
\label{tab:architecture}
\begin{ruledtabular}
\begin{tabular}{l c}
\textbf{Parameter} & \textbf{Value} \\
\colrule
\multicolumn{2}{c}{\textit{Architecture}} \\
Input dimension & 6 ($t, r, \phi, \dot t, \dot r, \dot \phi$) \\
Hidden layers & 4 \\
Neurons per layer & 500 \\
Activation function & GELU \\
Output & Scalar $L_\theta$ (1) \\
\\
\multicolumn{2}{c}{\textit{Training}} \\
Optimizer & RAdam \\
Weight decay & $1\times10^{-6}$ \\
Learning rate & $5\times10^{-4}$ \\
LR schedule & Cosine annealing, $T_{\max} = \text{epochs}/10$ \\
Minimum learning rate $\eta_{\min}$ & $5\times10^{-8}$ \\
Batch size & 128 \\
Epochs & 100 (best validation at epoch~91) \\
Training samples & 120\,000 \\
Train/validation split & 80/20 \\
\\
\multicolumn{2}{c}{\textit{Regularization}} \\
Sylvester penalty weight $\lambda_{\text{syl}}$ & 5 \\
Gradient clip norm & 1.0 \\
\end{tabular}
\end{ruledtabular}
\end{table}

\subsection{Dataset Variations and Generation parameters:}

The candidate-state construction and physical filtering procedures are defined in \cref{sec:sampling}, while the alternative direct-energy sampler is described in \cref{sec:energy_sampling}. Here, we specify the five dataset variants used to evaluate the effects of phase-space sampling:

\begin{enumerate}
    \item the $\emph{baseline sampler}$, which uses the original unfiltered random-state sampling procedure;
    \item the $\emph{energy sampler}$, which samples the conserved energy directly;
    \item the $\emph{constrained velocity sampler}$, which applies the velocity-based procedure with $(E_{\max}=1)$;
    \item the $\emph{relaxed-energy velocity sampler}$, which increases the cutoff to $(E_{\max}=1.001)$;
    \item the $\emph{mixed-velocity sampler}$, which combines narrow and broad radial-velocity proposal windows to increase the representation of both circular and more eccentric states.
\end{enumerate}

Baseline sampler relies on random selection of coordinates from the predetermined range list from Table.~\ref{tab:dataset_params} and used for benchmarking the performance of the other sampling methods.
3 out of the 5 datasets outlined in \cref{tab:dataset_params} uses the energy-constrined sampling methods that utilizes sampling methods outlined in \cref{sec:sampling} and \cref{sec:energy_sampling}

\begin{figure*}[tp]
    \centering
    % Top plot
    \includegraphics[width=\textwidth]{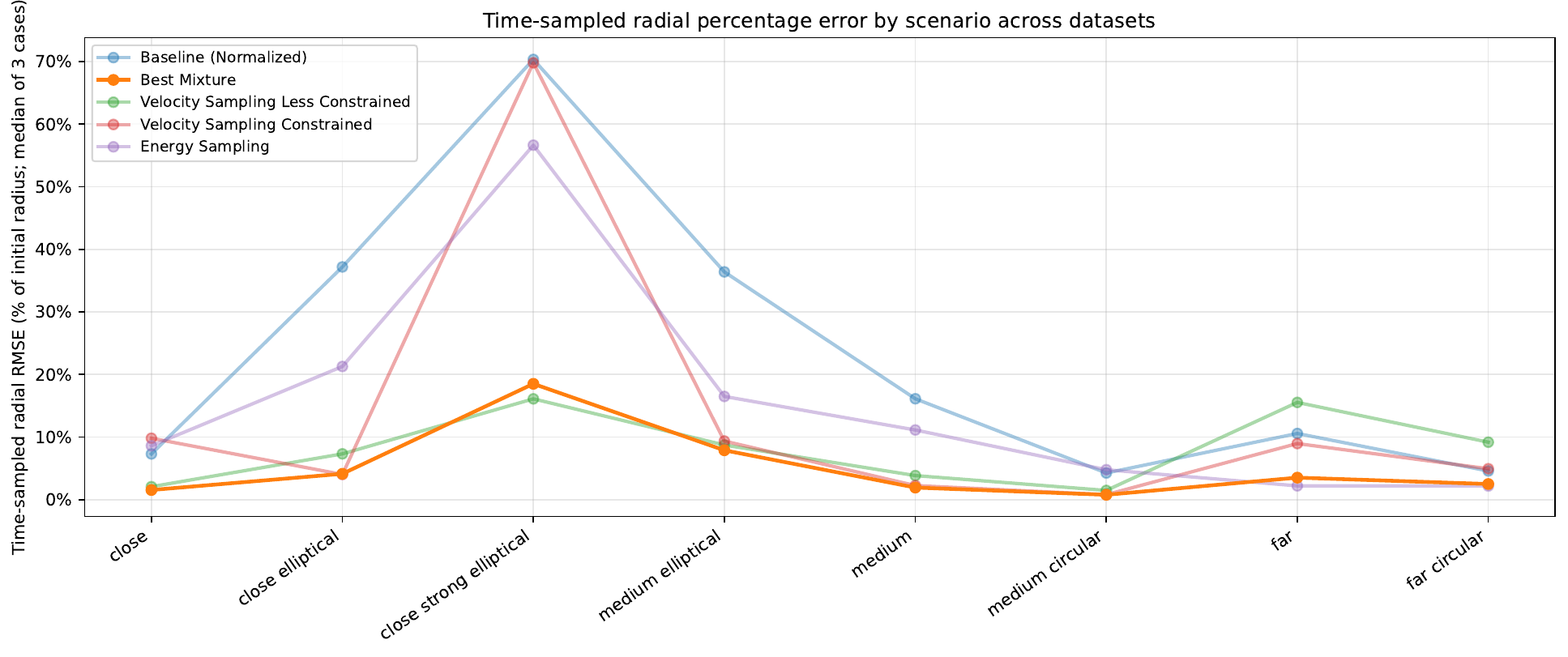}
     \caption{Mean radial error for each scenario and model across the test time span $0\leq\tau/M\leq5\times10^4$. 
            The mixed-velocity sampler (orange) yields the best trajectory prediction accuracy across all eight test scenarios, while the baseline s    ampler performs the worst. The energy-sampling method performs better than the baseline but worse than the velocity-based samplers at highly elliptical or far scenarios.} 
    \label{fig:traj_accuracy}
    \vspace{1em}

    % Bottom 3 panels
    \begin{minipage}[b]{0.32\textwidth}
        \centering
        \includegraphics[width=\textwidth]{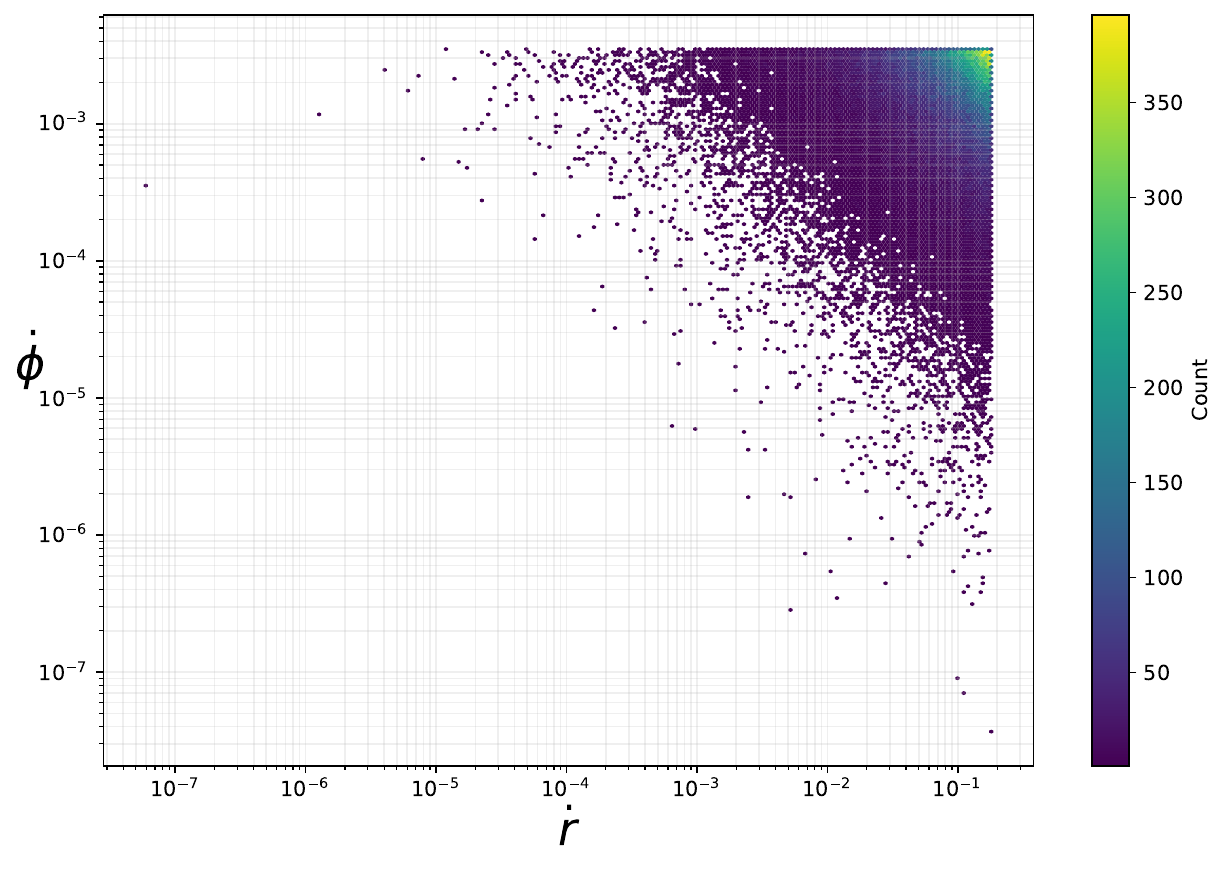}\\
        \small (a) Baseline
    \end{minipage}\hfill
    \begin{minipage}[b]{0.32\textwidth}
        \centering
        \includegraphics[width=\textwidth]{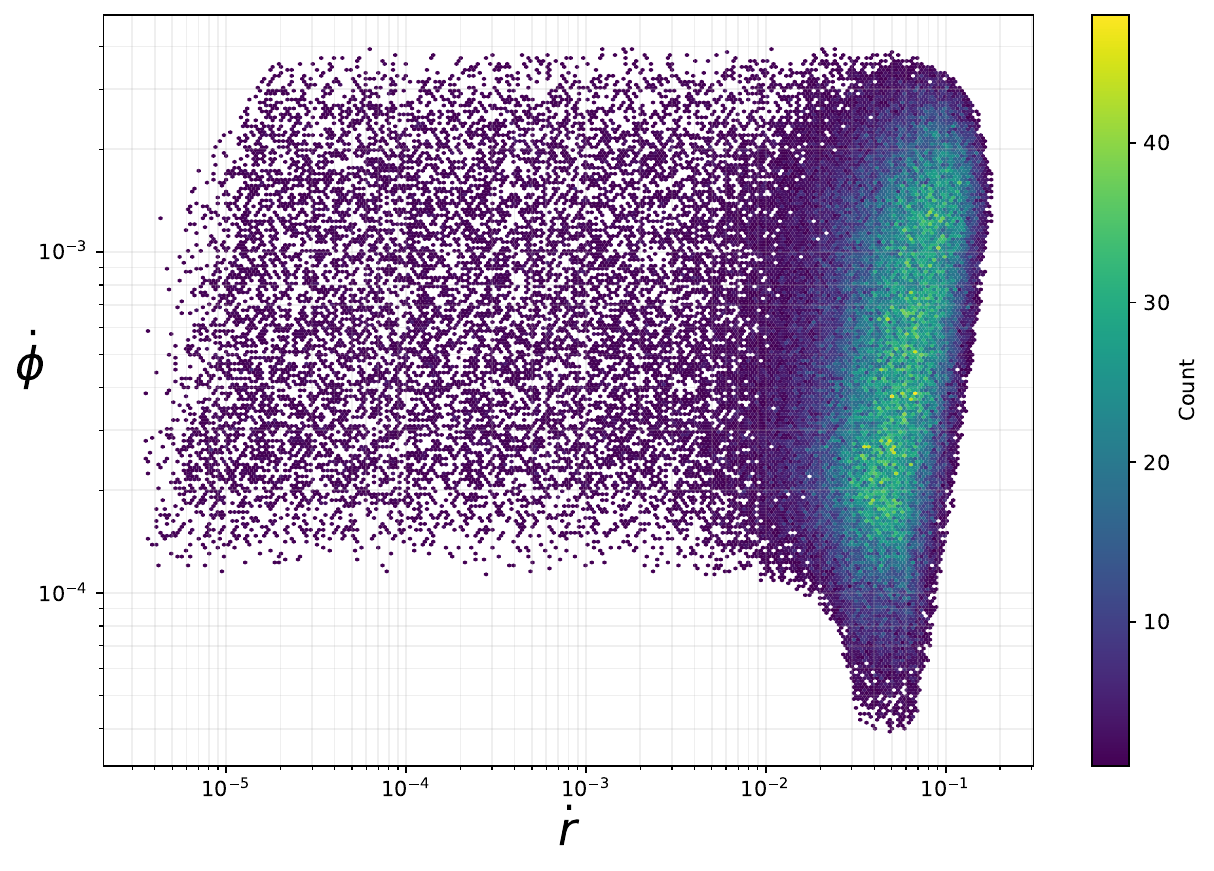}\\
        \small (b) Energy Sampling
    \end{minipage}\hfill
    \begin{minipage}[b]{0.32\textwidth}
        \centering
        \includegraphics[width=\textwidth]{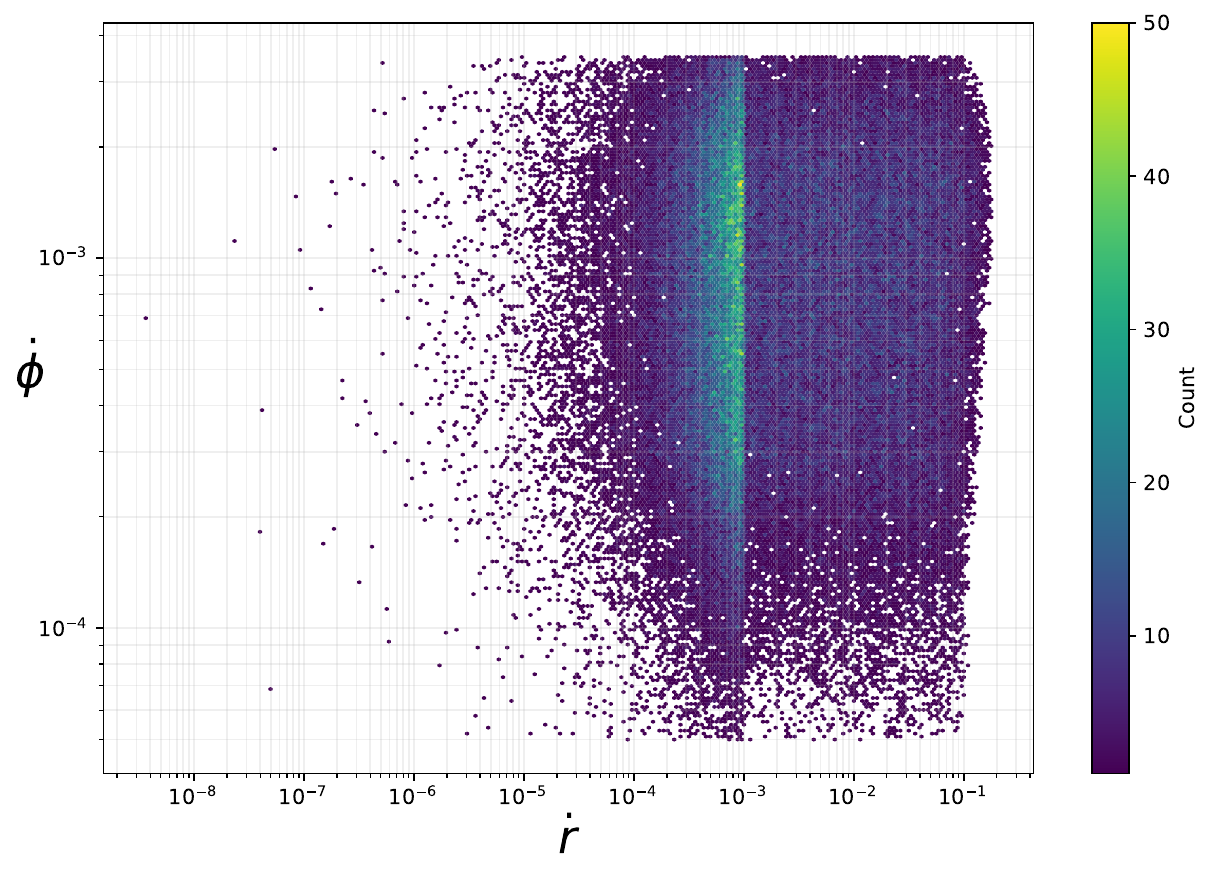}\\
        \small (c) Velocity Sampling (Best Mixture)
    \end{minipage}
     \caption{Velocity-space distributions for the baseline, energy, and
            mixed-velocity samplers listed in \cref{tab:dataset_params}. The panels show
            the density of samples in the \((\dot r,\dot\phi)\) plane and are not normalized
            for direct comparison.} 
    \label{fig:dists}
\end{figure*}

When we compare the energy and velocity sampling methods in terms of trajectory prediction accuracy, we see that velocity sampling yields better training performance, especially when the energy boundedness constrained is relaxed (\cref{fig:traj_accuracy}).
Based on this result, we trained our best model on a dataset generated by velocity sampling method an uses two sampling windows in radial velocity ranges:$ \{0, \num{1e-3} \} + \{0,\num{1.8e-1}\}$ sampled individually and then combined, balancing the distribution in velocity space.

For all the models, training is performed on the samples in between $r \in [\num{50}, \num{300}]$ units region where $r_s=\num{2}$ units (with $G=c=\num{1}$).
System has time and azimuthal angle $\phi$ symmetry thus, they are assigned randomly to each sample in the ranges $t \in \qtyrange{0}{1e5}{\second}$ and $\phi \in [0, 2\pi]\,\unit{\radian}$.
Each dataset contains \num{120000} sample points, \qty{20}{\percent} of which is used for validation. And the training is performed under the normalization described in \cref{sec:methods}.

In \cref{fig:dists}, we also present the data distributions and densities in spatial velocity-phase space to visualize the effectiveness of different sampling methods.
We observe that while the range of baseline sampling is sufficiently large, its sample densities are heavily biased towards high velocities.
And best mixture dataset offers the most uniform density/range ratio and thus the best training performance.

\begin{table*}[htbp]
    \caption{Velocity ranges, energy-filter thresholds, effective-potential
            ranges, and fractions of samples satisfying $(E<1)$ for the five dataset
            variants. All quantities are expressed in geometric units with $(G=c=1)$.}
    \label{tab:dataset_params}
    \begin{ruledtabular}
        \begin{tabular}{l c c c c c}
            \textbf{Dataset}
            & \textbf{$\dot{r}$ range ($\times \pm 1$ units/s)}
            & \textbf{$\dot{\phi}$ range ($\times \pm 10^{-3}$ rad)}
            & \textbf{$E_{\text{max}}$}
            & \textbf{$U_{\text{eff}}^2$ range}
            & \textbf{Bound Ratio} \\
            \colrule
            Baseline Sampling
            & $[\num{5.96046e-8}, \num{0.18000}]$
            & $[\num{3.67872e-5}, \num{3.49999}]$
            & --
            & $[\num{0.96002}, \num{2.08132}]$
            & \qty{27}{\percent} \\

            Energy Sampling
            & $[\num{3.56960e-6}, \num{0.17873}]$
            & $[\num{3.92156e-2}, \num{3.93016}]$
            & \num{0.999}
            & $[\num{0.96466}, \num{0.99800}]$
            & \qty{100}{\percent} \\

            Constrained Velocity Sampling
            & $[\num{2.00067e-3}, \num{0.17999}]$
            & $[\num{8.00204e-2}, \num{3.99858}]$
            & \num{1.000}
            & $[\num{0.96474}, \num{0.99999}]$
            & \qty{100}{\percent} \\

            Velocity Sampling
            & $[\num{1.61809e-6}, \num{0.17971}]$
            & $[\num{5.00841e-2}, \num{3.49992}]$
            & \num{1.001}
            & $[\num{0.96466}, \num{1.00198}]$
            & \qty{88}{\percent} \\

            \textbf{Best Mixture}
            & $[\num{3.64801e-9}, \num{0.17820}]$
            & $[\num{5.00190e-2}, \num{3.49994}]$
            & \num{1.001}
            & $[\num{0.96474}, \num{1.00200}]$
            & \qty{92}{\percent} \\

            \multicolumn{6}{l}{\footnotesize
            (The baseline sampler has no energy cutoff; its maximum observed energy is \(E=1.443\).)}
        \end{tabular}
    \end{ruledtabular}
\end{table*}

\FloatBarrier

\subsection{Performance Comparison of the Datasets:}

Here we present the performance metrics of the models trained on the different datasets in terms of loss convergence and mean trajectory prediction accuracy.

\begin{figure}[htbp]
    \centering
    \includegraphics[width=\columnwidth]{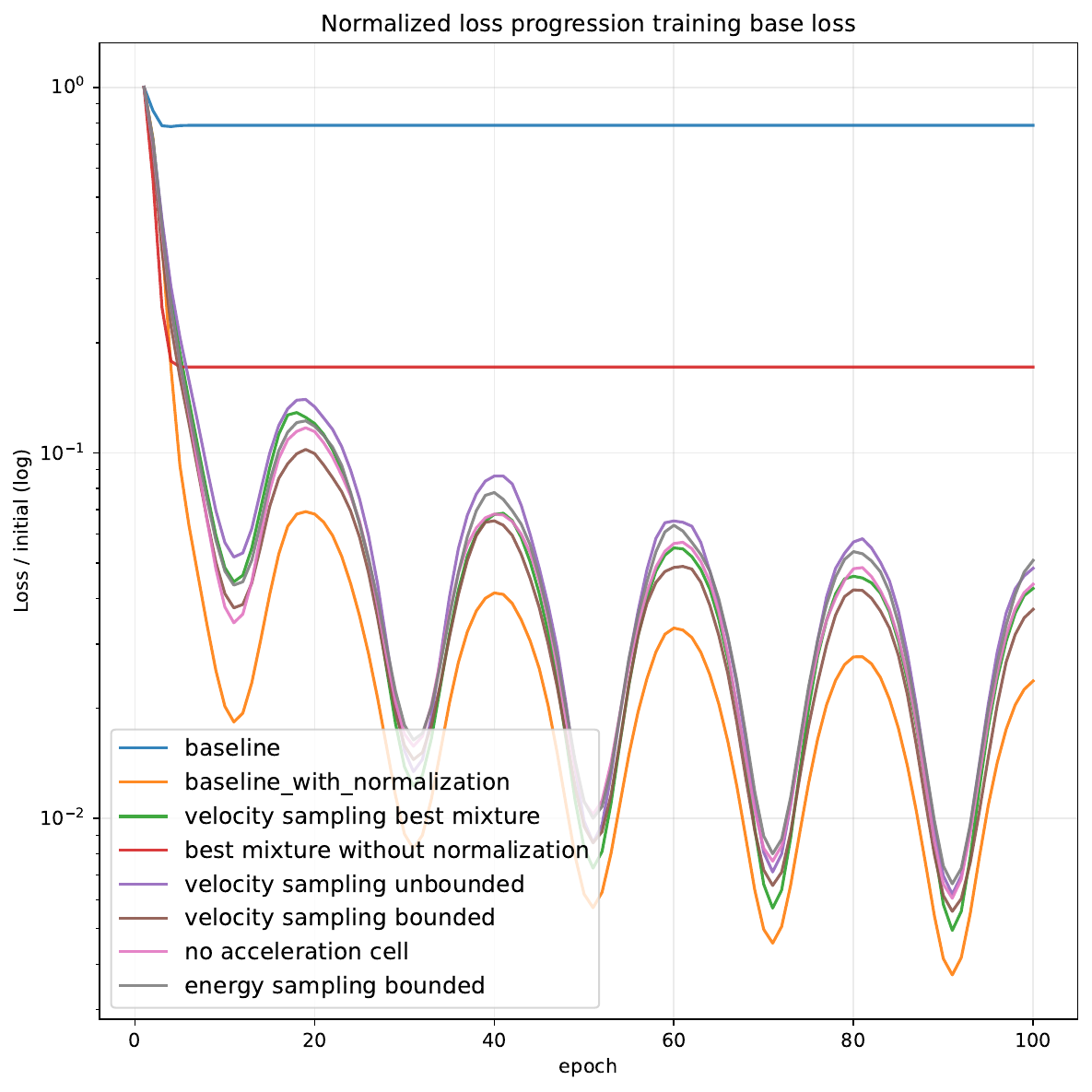}
    \caption{Progression of loss graph for baseline models and models trained with datasets created by energy sampling and velocity sampling. Y axis represents the ratio between the loss value in current epoch and the initial loss.}
    \label{fig:losses_all}
\end{figure}

\cref{fig:losses_all} shows the progression of training MAE loss for different datasets.
 The baseline sampler exhibits the largest fractional reduction in training
loss, but this ranking does not agree with the trajectory-level evaluation.
This difference demonstrates that relative single-step loss convergence alone
is not a reliable indicator of long-horizon rollout accuracy (See \cref{fig:norm_effects} and \cref{fig:traj_accuracy}).

Both models use the same decoupled normalization and the same per-coordinate loss weighting. Therefore, the performance gap is not caused by normalization itself, but by the training-state distribution induced by the dataset design.

A likely explanation is that the baseline sampler assigns substantially more weight to high-velocity and $(E\geq1)$ states, whereas the mixed-velocity
sampler increases the representation of low $\lvert\dot r\rvert$ and $(E<1)$ configurations. The training objective is therefore averaged over
different state distributions, and its fractional decrease need not reflect accuracy in the bound-orbit regime used for trajectory evaluation.

The mixed velocity dataset increases coverage of bounded states and near-circular/low-$\dot r$ regimes, which are harder to fit pointwise but are critical for stable integration. This numerical "tradeoff" allows the model to utilize most out of the decoupled normalization which is designed to prevent smaller numerical channels from being suppressed during optimization.   
Hence training with mixed velocity dataset, we observe worse average training loss yet better  delivery in terms of trajectory fidelity.

From \cref{fig:traj_accuracy}, we see that the mixed-velocity sampler yields the best trajectory prediction accuracy across all three orbital regimes, while the baseline sampler performs the worst, which also supports the dataset quality claim made above. The energy-sampling method performs better than the baseline but worse than the velocity-based samplers, indicating that direct energy sampling is less effective at capturing the relevant dynamics for long-horizon predictions.

\section{Alternative Dataset Sampling Methods and Energy Definitions}
\label{sec:dat}
\subsection{Killing Constants}\label{sec:killing}
Before we outline energy sampling method, let us define the killing constants which serves as constants of motion in this spacetime.Schwarzschild spacetime possesses two Killing vectors arising from its symmetries:
\textbf{Timelike Killing vector} $\xi^\mu = (1, 0, 0, 0)$ --- associated with time-translation symmetry, yielding conservation of specific energy $E$ along geodesics.
And \textbf{Rotational Killing vector} $\eta^\mu = (0, 0, 0, 1)$ --- associated with azimuthal symmetry, yielding conservation of specific angular momentum $L_z$ along geodesics.

For a geodesic with four-velocity $u^\mu = \dot{x}^\mu = dx^\mu/d\tau$, the conserved quantity associated with a Killing vector $K^\mu$ is $K_\mu u^\mu = \text{constant}$.

Using $\xi^\mu = \delta^\mu_t$:

\begin{align}
    \label{eq:energy_relation}
    E &\equiv -\xi_\mu u^\mu = -g_{\mu\nu}\xi^\nu u^\mu \nonumber \\ 
&= -g_{\mu t}\xi^t u^\mu = -g_{tt}\xi^t u^t \nonumber \\
    &= -\left[-\left(1-\frac{r_s}{r}\right)\right] \cdot 1 \cdot \dot{t} \nonumber \\
    E &= \left(1-\frac{r_s}{r}\right)\dot{t} 
\end{align}

And using $\eta^\mu = \delta^\mu_\phi$, on the equatorial plane ($\theta = \pi/2$, $\sin\theta = 1$):

\begin{align}
    \label{eq:momentum_relation}
    L &\equiv \eta_\mu u^\mu = g_{\mu\nu}\eta^\nu u^\mu \nonumber \\
    &= g_{\mu\phi}\eta^\phi u^\mu = g_{\phi\phi}\eta^\phi u^\phi \nonumber \\
    &= r^2\sin^2\theta \cdot 1 \cdot \dot{\phi} \nonumber \\
    L &= r^2\dot{\phi}
\end{align}

\subsection{Specific Energy Definition and Sampling method:}
\label{sec:energy_sampling}
\subsubsection{Specific Energy Definition:}
A massive particle following a timelike geodesic satisfies the normalization condition $g_{\mu\nu}\dot{x}^\mu\dot{x}^\nu = -1$. For a particle moving in the equatorial plane ($\theta = \pi/2$), this condition can be expressed as:
\begin{equation}
    -\left(1-\frac{r_s}{r}\right)\dot{t}^2 + \left(1-\frac{r_s}{r}\right)^{-1}\dot{r}^2 + r^2\dot{\phi}^2 = -1
\end{equation}

Using the definitions of the conserved quantities $E$ and $L$ from Killing vectors, we can express $\dot{t}$ and $\dot{\phi}$ in terms of $E$ and $L$. After some simplifications, we arrive at the specific energy relation for a particle in Schwarzschild spacetime \cite{Lim:2024mkb}:

\begin{equation} 
    \label{eq:specific_energy}
    E^2 = \dot{r}^{2} + \left(1-\frac{r_s}{r}\right)\left(1+\frac{L^2}{r^2}\right)
\end{equation}

\begin{equation*}
    U_{eff}=\left(1-\frac{r_s}{r}\right)\left(1+\frac{L^2}{r^2}\right)
\end{equation*} 
Where $U_{eff}$ is called the effective potential of the particle based on its distance from the black hole and its angular momentum.

\subsubsection{Innermost Stable Circular Orbit:}

For that massive particles orbit to stable and circular the following conditions must hold:

\begin{equation*}
    E^2=U_{eff}, \text{ }
    U_{eff}' = 0,\text{ } U_{eff}''>0    
\end{equation*}
When we do the differentiation and set the second derivative to 0, we see the condition $r > 3r_s$ must hold. This result gives us the first hint for finding a lover bound to energies in our dataset and is explored in detail in appendix B.

Using the first derivative condition $U_{eff}'=0$ we can relate angular momentum of a particle to its radius and when we plug in $r=3r_s$ we obtain a lowermost angular momentum for a particle orbiting the innermost stable circular orbit is:
\begin{equation}
    \label{eq:ISCO}
    L_{ISCO}=\sqrt{3r_s^2} 
\end{equation}

\subsubsection{Sampling E Based on $L_{ISCO}$:}
One of the methods of generating a dataset using the ISCO lower limit for angular momentum is to sample energy states directly. We use the result from \eqref{eq:ISCO} in \eqref{eq:specific_energy} and set $\dot{r}=0$, we obtain a relation for the minimum required energy for an orbit to be stable around the black hole as a function of r where r represents the farthest point or the apoapsis of that orbit. We set the boundaries for the energies of our sample between this $E_{min}$ and 0.999 as $E=1$ would be an unbounded flyby orbit.

For our experiments we sample radius values log-uniformly from the range $r=50-300$ units where $r_s=2$ units. For a black hole of this size minimum angular momentum necessary for an innermost stable orbit is $L_{ISCO}=\sqrt{12}\approx3.464$. We then set the lower energy bound for each r using:
\begin{equation*}
    E_{min}^2 =  \left(1-\frac{2}{r}\right)\left(1+\frac{L_{ISCO}^2}{r^2}\right)
\end{equation*}
And finally assigning the total energy to each sample $(E_{sample})$ as a random value between the $E_{min}$ and 0.999 we cover the bounded orbit radius-energy phase space entirely. In order to create variety in eccentricities we distribute the difference between $E_{sample}$ and $E_{min}$ to the $\dot{r}$ and some $L_{additional}$ contributions to the total energy. 
\begin{gather*}
    E_{sample}^2-E_{min}^2=\Delta{} \\
    \left(1-\frac{2}{r}\right)\left(1+\frac{L^2}{r^2}\right)  =  \left(1-\frac{2}{r}\right)\left(1+\frac{L_{ISCO}^2}{r^2}\right) +\alpha\Delta{}\\
    \implies L^2 = L_{ISCO}^2+\frac{r^2}{1-2/r}\alpha\Delta{}\\
    \dot{r}^2=(1-\alpha)\Delta{}\\
\end{gather*}

During our experiments, we noticed model performed better at the task of predicting small precession orbits when for the $85\%$ of the samples in the dataset, $\alpha$ is uniform randomly chosen between 0-1 and for the remaining $15\%$ distribution is biased towards the lower $\dot{r}$ via log-uniform sampled $\alpha$ values between 1 and 0. 

Although this method of sampling offers a more uniform coverage of the radius-energy phase space, it is not as physically motivated as velocity sampling in terms of sample distribution. Moreover, performance comparisons between the two methods show that velocity sampling outperforms Energy sampling in terms of model accuracy and generalization. This is also visible in \cref{fig:losses_all} where veloicty sampled best mixture dataset is shown with green loss progression line. 

\section{Normalization, Euler--Lagrange Distortion, and Hessian De-normalization}
\label{sec:norm}

\subsection{Normalization map used in the methodology}

Let $q,\dot q,\ddot q \in \mathbb{R}^n$ denote physical coordinates, velocities, and accelerations.  
The training pipeline applies per-coordinate scaling:
\begin{align}
\tilde q_i &= S_i q_i, \\
\tilde{\dot q}_i &= \frac{S_i}{T_i}\dot q_i, \\
\tilde{\ddot q}_i &= \frac{S_i}{T_i^2}\ddot q_i.
\end{align}
Define diagonal matrices
\begin{equation}
D_q := \mathrm{diag}(S_1,\dots,S_n), 
\qquad
D_v := \mathrm{diag}\!\left(\frac{S_1}{T_1},\dots,\frac{S_n}{T_n}\right),
\end{equation}
so that
\begin{equation}
\tilde q = D_q q, \qquad \tilde v = D_v v,\quad (v\equiv \dot q).
\end{equation}

Thus the model is trained/evaluated in normalized state coordinates
\begin{equation}
x = \begin{bmatrix} q \\ v \end{bmatrix}, 
\qquad
\tilde x = \begin{bmatrix} \tilde q \\ \tilde v \end{bmatrix}
=
\begin{bmatrix} D_q & 0 \\ 0 & D_v \end{bmatrix}
\begin{bmatrix} q \\ v \end{bmatrix}.
\end{equation}

\subsection{Euler--Lagrange equation and distortion under anisotropic scaling}

In physical coordinates, write the Euler--Lagrange equation in index form:
\begin{equation}
\sum_j H_{ij}\ddot q_j + \sum_j C_{ij}\dot q_j = g_i,
\end{equation}
with
\begin{equation}
H_{ij} := \frac{\partial^2 L}{\partial \dot q_i \partial \dot q_j},\quad
C_{ij} := \frac{\partial^2 L}{\partial \dot q_i \partial q_j},\quad
g_i := \frac{\partial L}{\partial q_i}.
\end{equation}

Under $\tilde q_i=S_i q_i,\ \tilde{\dot q}_i=(S_i/T_i)\dot q_i$, the derivative scalings are
\begin{equation}
\frac{\partial}{\partial \dot q_i}
= \frac{S_i}{T_i}\frac{\partial}{\partial \tilde{\dot q}_i},
\qquad
\frac{\partial}{\partial q_i}
= S_i\frac{\partial}{\partial \tilde q_i},
\end{equation}
hence
\begin{align}
\tilde H_{ij}
:= \frac{\partial^2 L}{\partial \tilde{\dot q}_i \partial \tilde{\dot q}_j}
&= H_{ij}\frac{T_iT_j}{S_iS_j},
\\
\tilde C_{ij}
:= \frac{\partial^2 L}{\partial \tilde{\dot q}_i \partial \tilde q_j}
&= C_{ij}\frac{T_i}{S_iS_j},
\\
\tilde g_i
:= \frac{\partial L}{\partial \tilde q_i}
&= \frac{1}{S_i}g_i.
\end{align}
Or equivalently,
\begin{align}
H_{ij} &= \tilde H_{ij}\frac{S_iS_j}{T_iT_j}, \\
C_{ij} &= \tilde C_{ij}\frac{S_iS_j}{T_i}, \\
g_i &= S_i\tilde g_i.
\end{align}

Substituting into the physical equation and rewriting in normalized variables gives
\begin{equation}
\tilde{\ddot q}_k
=
\sum_i \tilde H^{-1}_{ki}\frac{T_i}{T_k}
\left[
\tilde g_i
-
\sum_j \tilde C_{ij}\frac{T_j}{T_i}\tilde{\dot q}_j
\right].
\label{eq:distorted_EL}
\end{equation}

\subsubsection{Interpretation of the $T_j/T_i$ factors}
Equation~\eqref{eq:distorted_EL} is obtained by starting from the physical Lagrangian $L_{\rm phys}(q,\dot q)$, substituting
$q=D_q^{-1}\tilde q$ and $\dot q=D_v^{-1}\tilde{\dot q}$, and rewriting the physical Euler-Lagrange equations in terms of the normalized variables.
The resulting $T_j/T_i$ factors arise because the position and velocity arguments are scaled independently. In particular, when the
$T_i$ are not all equal, $\tilde{\dot q}_i$ cannot generally be interpreted as the derivative of $\tilde q_i$ with respect to a single
common evolution parameter.

The model does not use the transformed physical Lagrangian or the corresponding operator in Eq.~\eqref{eq:distorted_EL}. Instead, it learns
a latent scalar $L_{\rm learned}(\tilde q,\tilde{\dot q})$ directly as a function of the normalized position and velocity features. 
The normalized acceleration map is then defined through the Euler--Lagrange-structured operator
\begin{equation}
\tilde{\ddot q} = \tilde H^{-1}\bigl[\tilde g - \tilde C \tilde{\dot q}\bigr],
\label{eq:EL_norm_standard}
\end{equation}
where 
\begin{equation}
\tilde H_{ij} = \frac{\partial^2 L_{\rm learned}}{\partial\tilde{\dot q}_i\partial\tilde{\dot q}_j}, \quad
\tilde C_{ij} = \frac{\partial^2 L_{\rm learned}}{\partial\tilde{\dot q}_i\partial\tilde q_j}, \quad
\tilde g_i = \frac{\partial L_{\rm learned}}{\partial\tilde q_i}.
\end{equation}

No $T_j/T_i$ factors appear in \cref{eq:EL_norm_standard} because this equation defines the operator used by the network which is not obtained by transforming the
physical Euler--Lagrange equations under the decoupled normalization. Accordingly, the learned scalar $L_{\rm learned}(\tilde q, \tilde{\dot q})$ is interpreted as a latent generating 
scalar for the normalized acceleration field, not as the physical Lagrangian expressed in normalized coordinates.

The $T_j/T_i$ factors in Eq.~\eqref{eq:distorted_EL} therefore characterize how the physical Euler--Lagrange equations would appear after the independently chosen position and velocity scalings. They are
not additional factors applied by the model. Since the training procedure does not impose
\begin{equation}
    L_{\rm learned}(\tilde q,\tilde{\dot q})
    =
    L_{\rm phys}
    \left(
        q(\tilde q),
        \dot q(\tilde{\dot q})
    \right),
\end{equation}
the two scalars need not have the same functional form. The deviations observed in the learned velocity Hessian (\cref{sec:hessian_chapter}) are consistent with the network having
constructed a distinct latent representation whose derivatives generate the target normalized acceleration field through Eq.~\eqref{eq:EL_norm_standard}.

\subsection{Hessian de-normalization for model-vs-analytic comparability}

The learned Lagrangian network receives normalized inputs $\tilde x$ and yields a scalar $L_\theta(\tilde x)$.  
Its velocity Hessian is therefore naturally computed in normalized velocity coordinates:
\begin{equation}
\tilde H^{(\theta)}_{vv}
=
\frac{\partial^2 L_\theta}{\partial \tilde v\,\partial \tilde v}.
\end{equation}
To compare with an analytical Hessian in physical coordinates,
\begin{equation}
H^{(a)}_{vv}
=
\frac{\partial^2 L_a}{\partial v\,\partial v},
\end{equation}
we must map $\tilde H^{(\theta)}_{vv}$ back to physical coordinates.

\paragraph{Chain-rule pullback.}
Since $\tilde v = D_v v$,
\begin{equation}
\frac{\partial}{\partial v}
=
D_v^\top \frac{\partial}{\partial \tilde v}
\quad\text{(for diagonal $D_v$, $D_v^\top=D_v$)}.
\end{equation}
Applying chain rule twice:
\begin{equation}
H^{(\theta)}_{vv,\text{pullback}}
=
D_v^\top \tilde H^{(\theta)}_{vv} D_v
=
D_v \tilde H^{(\theta)}_{vv} D_v.
\label{eq:hess_pullback_matrix}
\end{equation}
Componentwise:
\begin{equation}
\left(H^{(\theta)}_{vv,\text{pullback}}\right)_{ij}
=
\left(\frac{S_i}{T_i}\right)
\left(\frac{S_j}{T_j}\right)
\left(\tilde H^{(\theta)}_{vv}\right)_{ij}.
\label{eq:hess_pullback_component}
\end{equation}
\section{Additional Prediction Plots}
\label{sec:predictions}
This appendix presents additional trajectory prediction plots for all eight orbital scenarios across the close, medium, and far regions, alongside phase-space distribution analyses for the baseline and best-mixture datasets.

\begin{figure*}[htbp]
    \centering
    \includegraphics[width=\textwidth]{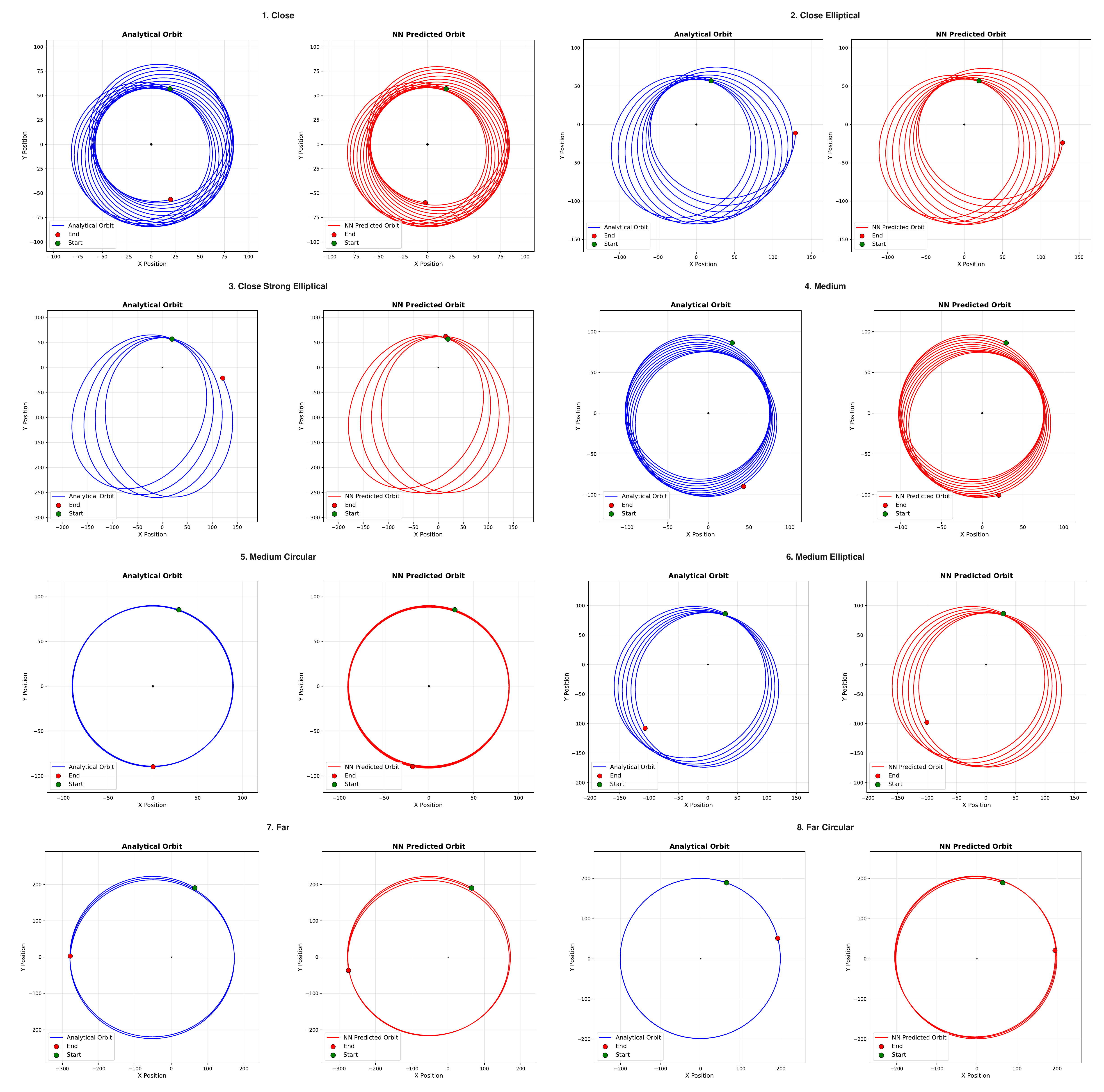}
    \caption{Trajectory predictions for all eight orbital scenarios arranged in a $4 \times 2$ grid across the close (1--3), medium (4--6), and far (7--8) regions. Each scenario block presents a side-by-side comparison between the analytical geodesic (left) and the neural network prediction (right).}
    \label{fig:all_scenarios_grid}
\end{figure*}

\begin{figure*}[htbp]
    \centering
    \includegraphics[width=\textwidth]{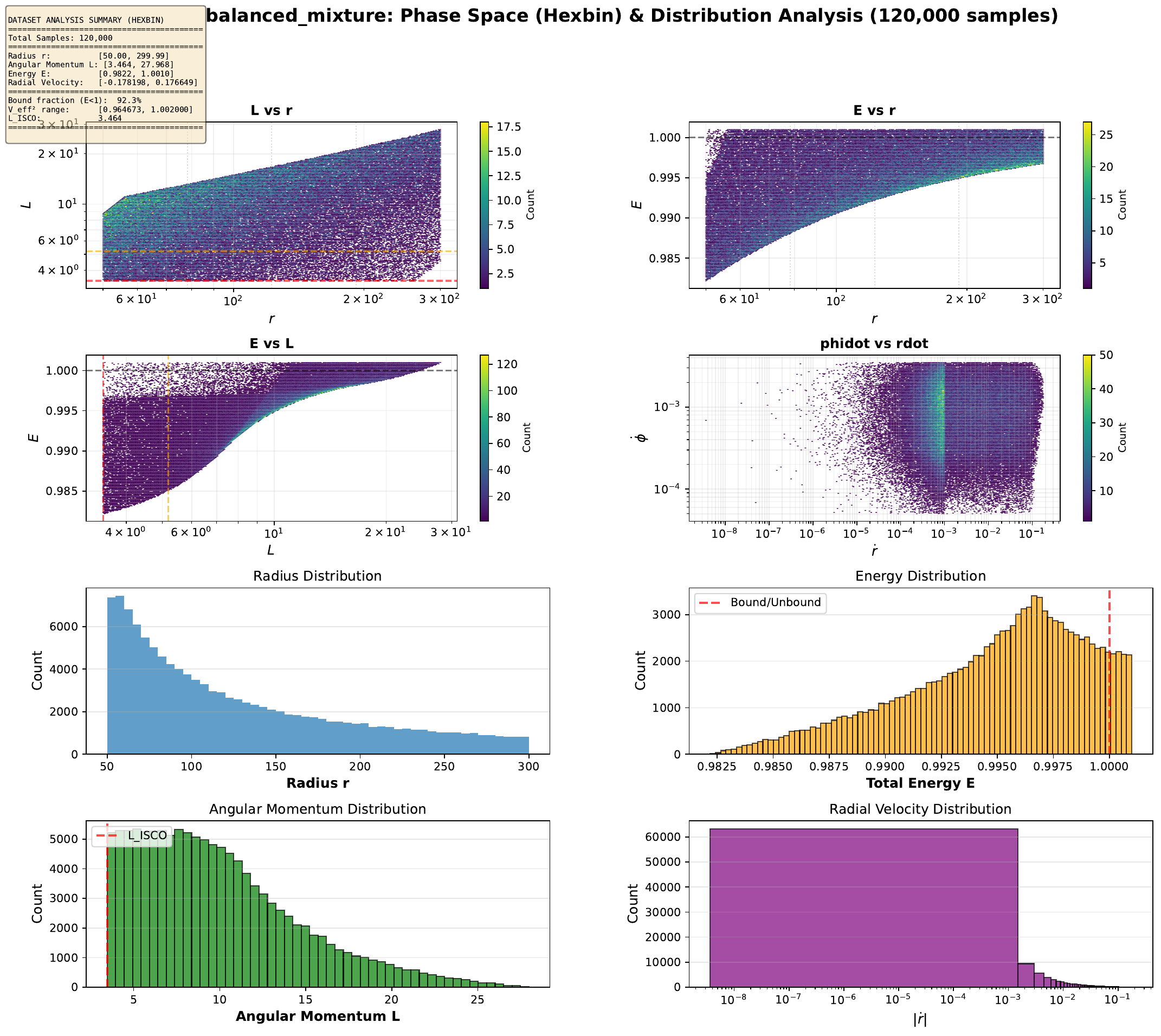}
    \caption{Hexbin density analysis of the best-mixture (velocity sampling) dataset in the $(\dot{r}, \dot{\phi})$ velocity phase-space.
    The balanced mixture combines two radial velocity windows to achieve the most uniform coverage of bound-orbit phase-space.}
    \label{fig:balanced_mixture_hexbin}
\end{figure*}

\begin{figure*}[htbp]
    \centering
    \includegraphics[width=\textwidth]{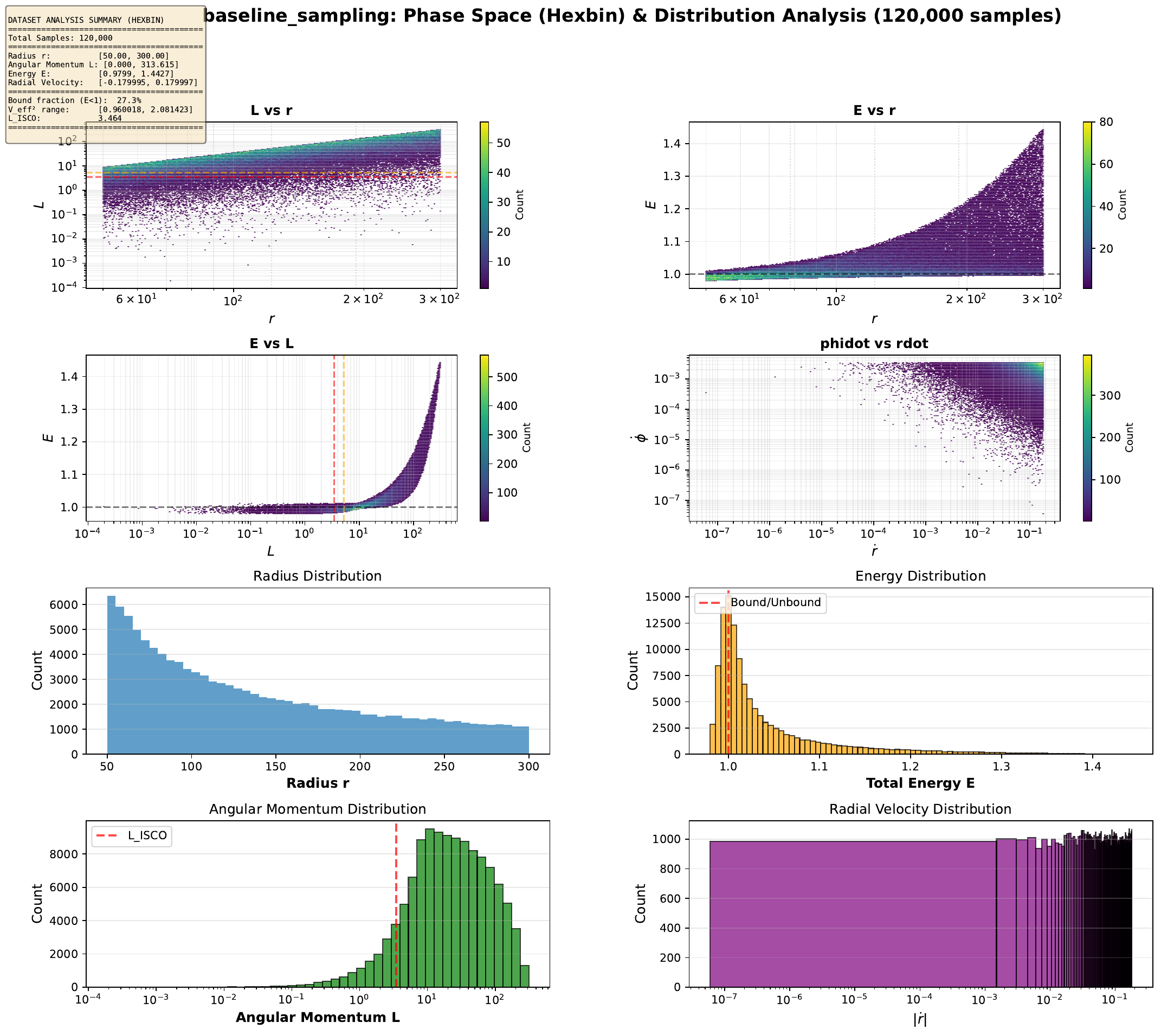}
    \caption{Hexbin density analysis of the baseline sampling dataset in the $(\dot{r}, \dot{\phi})$ velocity phase-space.
    The baseline dataset exhibits strong density bias toward high-velocity regions and contains only \qty{27.3}{\percent} bound orbits, compared to \qty{92.3}{\percent} for the best-mixture dataset.}
    \label{fig:baseline_sampling_hexbin}
\end{figure*}

\end{document}